\pdfoutput=1
\RequirePackage{fix-cm}

\documentclass[10pt,twocolumn,letterpaper]{article}

\usepackage[pagenumbers]{cvpr}

\usepackage{graphicx}
\usepackage{amsmath}
\usepackage{amssymb}
\usepackage{booktabs}
\usepackage{multirow}

\usepackage{cuted}
\usepackage{caption}

\newcommand{\method}{\textup{\textsc{StateStream}}}

\usepackage{algorithm}
\usepackage{algpseudocode}

\usepackage{tcolorbox}
\usepackage{needspace}
\tcbuselibrary{listings,breakable}
\newtcblisting{agentprompt}[2][]{
    title={#2},
    title after break={#2\space(continued)},
    colback=gray!5,
    colframe=gray!50!black,
    coltitle=white,
    fonttitle=\small\bfseries,
    boxrule=0.45pt,
    arc=1mm,
    left=2mm, right=2mm, top=1.5mm, bottom=1.5mm,
    before skip=7pt, after skip=8pt,
    breakable,
    lines before break=4,
    listing only,
    listing options={
        basicstyle=\fontencoding{T1}\fontfamily{lmtt}\fontsize{8pt}{8.8pt}\selectfont,
        columns=fullflexible,
        keepspaces=true,
        showstringspaces=false,
        breaklines=true,
        breakatwhitespace=true,
        breakautoindent=false,
        breakindent=1em,
        upquote=true,
        tabsize=2,
        aboveskip=0pt,
        belowskip=0pt
    },
    #1
}

\definecolor{cvprblue}{rgb}{0.21,0.49,0.74}
\usepackage[pagebackref,breaklinks,colorlinks,allcolors=cvprblue,hypertexnames=false]{hyperref}

\usepackage{booktabs}

\usepackage{colortbl}
\usepackage{tabularx}
\definecolor{perfblue}{RGB}{70,130,180}
\definecolor{pageviolet}{RGB}{122,90,248}

\usepackage{graphicx}

\newcommand{\metricgain}[2]{%
  \textsuperscript{\textcolor{#1}{\scriptsize #2}}%
}

\title{StateStream: Agentic World-State Reasoning and Control\\for Interactive Long Video Generation}

\author{Junwei Zhou \qquad Yuyao Zhang \qquad Yu-Wing Tai\\[8pt]
Dartmouth College\\[6.4pt]
{\normalsize\itshape\href{https://colezwhy.github.io/statestream-page/}{\textcolor{pageviolet}{colezwhy.github.io/statestream-page}}}}

\hypersetup{
  pdftitle={StateStream: Agentic World-State Reasoning and Control for Interactive Long Video Generation},
  pdfauthor={Junwei Zhou, Yuyao Zhang, Yu-Wing Tai}
}

\makeatletter
\patchcmd{\@maketitle}{\vspace*{12pt}}{\vspace*{0pt}}{}{}
\patchcmd{\@maketitle}{\vspace*{24pt}}{\vspace*{19pt}}{}{}
\patchcmd{\@maketitle}{\vskip .5em}{\vskip 0pt}{}{}
\patchcmd{\@maketitle}{\vskip .375in}{\vskip .2in}{}{}
\makeatother

\begin{document}
\maketitle
      \begingroup
\setlength{\stripsep}{0pt}
\begin{strip}
    \centering
    \includegraphics[width=\textwidth]{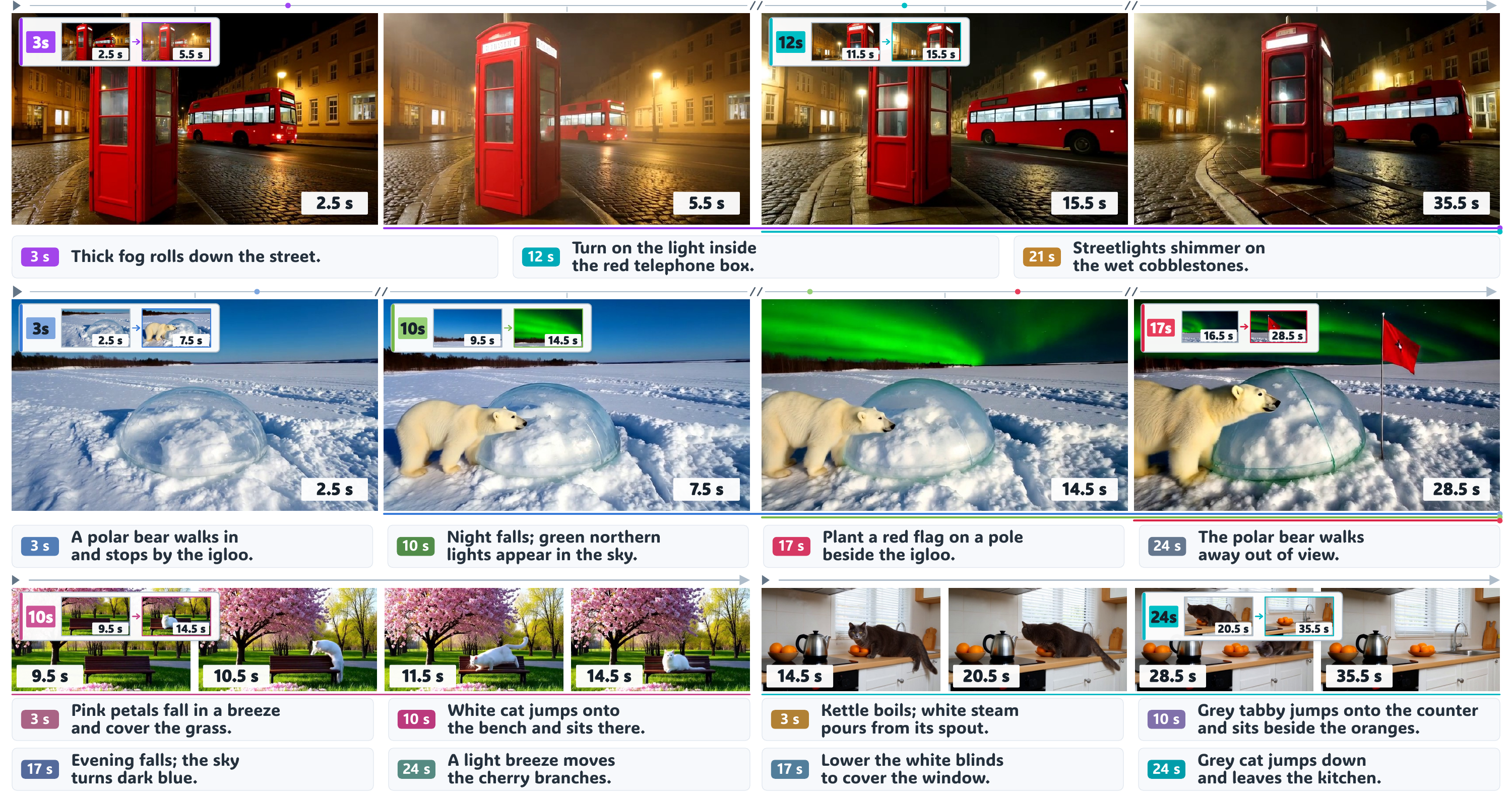}
    \captionsetup{hypcap=false}
    \captionof{figure}{\textbf{Interactive long video generation with persistent world state.} \method{} follows successive interactive prompts across four scenes, combining environmental changes, local object updates, and entity arrivals and departures while retaining changes that later prompts do not supersede. Prompts and their arrival times appear below the sequences; timestamps mark sampled frames, and colored insets compare local appearance before and after selected interactions.}
    \captionsetup{hypcap=true}
    \label{fig:teaser}
    \vspace{6pt}
\end{strip}
\endgroup

\begin{figure*}[t]
    \centering
    \includegraphics[width=\textwidth]{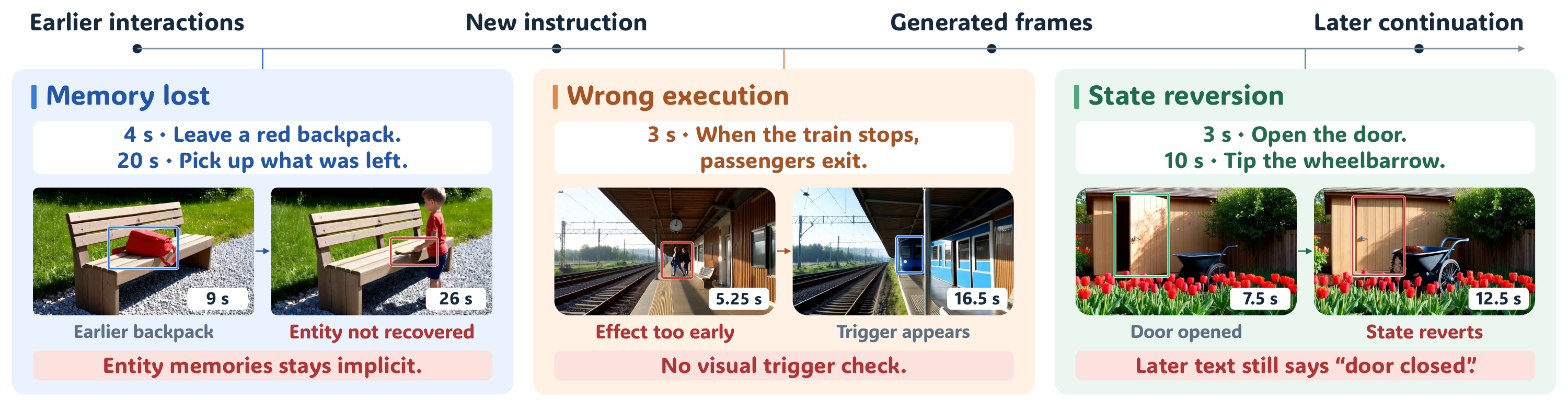}
    \caption{\textbf{Observed failures in existing video generators.}
    Left (LongLive): does not keep contents under a memory-dependent prompt.
    Middle (LongLive): executes interactions out of order, without a visual trigger check.
    Right (TempAct): reverts the drawn interactions without request.
    Red annotations highlight failed outcomes and the conditioning or control limitations.}
    \label{fig:motivation}
\end{figure*}

{\renewcommand{\endabstract}{\par}\begin{abstract}
Generating interactive long videos requires tracking an evolving world and reasoning about the consequences of user interactions, two essential capabilities for world modeling.
Recent work emphasizes visual consistency over extended durations, yet state-aware interactivity also requires interpreting interactive prompts in context and retaining their effects through the subsequent generation process.
To address this challenge, we present \method{}, an agentic framework for world-state reasoning and planning on a frozen video generator without task-specific training.
The framework links prompt interpretation and visual evidence through a shared world-state representation.
Specifically, \emph{World-State Planning} uses a persistent state program to resolve memory-dependent prompts and construct execution plans with action dependencies.
\emph{State-Guided Memory Control} coordinates persistent text conditioning with state-guided access to and local revision of entity-indexed visual references.
\emph{Feedback-Guided Execution} checks generated frames to govern step progression and memory updates, distinguishing requested changes from accepted visual evidence.
We also introduce an evaluation protocol using timed interactive prompts and visual outcome checks to assess execution and long-horizon state retention, with causal and conditional tasks alongside paired interaction evaluation.
Experiments show improved interaction success and long-horizon state retention, with gains on causal and conditional tasks and comparable visual quality, marking a step toward state-aware world modeling.

\end{abstract}
}
\raggedbottom
\section{Introduction}
\label{sec:intro}

Interactive long video generation requires the generated world to respond coherently to users as it evolves. Recent autoregressive models have advanced streaming synthesis and long-horizon visual consistency~\cite{selfforcing,longlive}. Interactive prompts let users introduce entities, alter environments, and request actions grounded in earlier events. These interactions require capabilities central to world modeling: reasoning about the evolving scene, realizing requested changes, and retaining consequences until superseded.

The relevant \emph{world state} includes entities, their attributes and locations, and global conditions. In Fig.~\ref{fig:teaser}, the bear remains present as an aurora and a flag are introduced, illustrating \emph{state retention}: keeping realized outcomes valid through subsequent generation. A hypothetical request to \emph{bring back an animal that previously left} would be \emph{memory-dependent}, requiring information from earlier interactions. Likewise, a request to \emph{open the door after the bear reaches the igloo} would be \emph{conditional}, requiring an observed trigger. Together, these requirements determine what to generate, when to act, and what must persist.

Existing approaches support prompt switching~\cite{longlive}, temporal decomposition~\cite{tempact}, entity-aware memory~\cite{iamflow,slotmemory}, and explicit state reasoning~\cite{stateagent}. Connecting these decisions to visual generation remains challenging. Retained references may reinforce outdated states, while refreshed references may contain incomplete outcomes; a planned transition may never appear. Figure~\ref{fig:motivation} illustrates failures in recovering an earlier entity, respecting a trigger, and retaining a realized change. Addressing these problems requires coordinating intended transitions, the visual evidence used for generation, and confirmation of their execution.

We introduce \method{}, an agentic framework that connects world-state reasoning, generation control, and visual feedback. A persistent \emph{state program} records entities and requested transitions, preserving their history. Shared entity identities link this program to conditioning text and visual references, while \emph{execution state} tracks progress and confirmation. Three modules implement this connection on a frozen video generator without task-specific training:

\textbf{World-State Planning} interprets memory-dependent prompts and state-dependent branches using the state program. It produces state updates and execution plans with actions, intended results, and visual checks; waiting steps represent conditions to observe before acting.
\textbf{State-Guided Memory Control} combines persistent scene descriptions (World-State text) with the active step description, and coordinates entity-indexed visual references. Text-Guided KV Extraction selects local generator features, while Align \& Read aligns stored references and guides their state-dependent use.
\textbf{Feedback-Guided Execution} checks generated frames to control step progression and confirm results. New local evidence updates persistent memory only after the relevant result is confirmed. Unconfirmed outcomes retain their target descriptions and are rechecked.

Evaluation covers execution, state retention, and visual quality, including causal and conditional tasks and paired interaction evaluation with identical target prompts and different histories. Experiments show improved interaction success and long-horizon retention over video baselines, with and without further prompts, and visual quality close to the backbone. Ablations support visual memory and targeted improvements in conditional timing from planning and feedback; a second frozen backbone supports transfer within autoregressive generation.

We summarize our contributions as follows:
\begin{itemize}
\item We introduce \method{}, connecting memory-dependent reasoning, execution planning, and interactive video generation through shared world state.
\item We develop state-guided memory control that links state transitions to entity-reference selection and alignment, with local memory updates gated by visual confirmation.
\item We establish an evaluation protocol for interactive long
video generation. Comprehensive experimental results demonstrate that our \method{} outperforms previous state-of-the-art methods, producing vivid long videos with reasonable interactions.
\end{itemize}

\section{Related Work}
\label{sec:related}

\subsection{Interactive Long Video Generation}
Diffusion-based sequence models support flexible temporal conditioning and long rollouts by denoising frames at different noise levels~\cite{diffusionforcing,historyguidance,rollingforcing}. Autoregressive distillation from bidirectional video models~\cite{wan} and rollout-aware training further reduce sampling latency and the mismatch between training and streaming inference~\cite{causvid,selfforcing,causalforcing,longlive}. 
Interactive systems accommodate changing prompts through conditioning updates, cache adaptation, and persistent entity references~\cite{phenaki,ditctrl,longlive,infinityrope,headforcing,iamflow,slotmemory}. 
Recent world models condition scene evolution on actions and camera motion~\cite{genie,gamengen,contextasmemory,worldmem,worldplay}, while temporal planners decompose complex instructions and coordinate execution~\cite{tempact}. \method{} uses a shared world-state representation to connect the interpretation of memory-dependent prompts with generation control. Feedback-Guided Execution checks generated frames to guide step progression and authorize local memory updates.

\subsection{World State and Visual Memory}
Retrieval-based memory extends the recent-frame window by selecting relevant historical frames, latents, or KV features as context for continued generation~\cite{contextasmemory,worldmem,longliverag,omnimem,memflow}. Complementary approaches use context retention, compression, or positional-encoding adaptation to sustain long rollouts under bounded memory budgets~\cite{streamingllm,contextforcing,deepforcing,headforcing,memrope,infinityrope}. The influence of distant references has also been studied in different aspects~\cite{lol,recencyforcing,groundedforcing}. Entity-aware approaches organize visual memory or predicted scene states around object identities and attributes to support consistency and targeted changes~\cite{iamflow,slotmemory,stateagent}. Our State-Guided Memory Control coordinates World-State text with entity-indexed visual references through Text-Guided KV Extraction and Align \& Read. Entity identities link requested updates to reference selection and placement, while State Verification confirms relevant outcomes before new local evidence is written to memory.

\subsection{Programs and Planning for Generative Worlds}
Explicit layouts and structured scene states provide controllable interfaces for constructing and refining compositional 3D scenes~\cite{holodeck2,layoutvlm,layoutyour3d2025zhou,zhou2026perceivethenplanlayoutaspolicymonocular3d,agentic3dscene}. Video planning methods organize narrative descriptions into scene sequences, shot specifications, or action stages that guide synthesis~\cite{videodirectorgpt,lvd,movieagent,filmagent,tempact}. Agentic scene-generation systems combine spatial planning with visual guidance and feedback to refine layouts, object composition, and physical plausibility~\cite{holodeck2,worldcraft,scenethesis,sage,sld}.
Executable scene programs provide interfaces for staged construction and targeted editing of structured environments~\cite{code4scene}. Programmable world models separate state evolution from visual realization through executable events and transition rules~\cite{cwm,coder,pwm}. \method{} connects World-State Planning to conditioning text and visual memory through shared entity identities. Feedback-Guided Execution checks generated frames to guide step progression and authorize local memory updates, keeping requested transitions from confirmed outcomes.

\begin{figure*}[t]
    \centering
    \includegraphics[width=\textwidth]{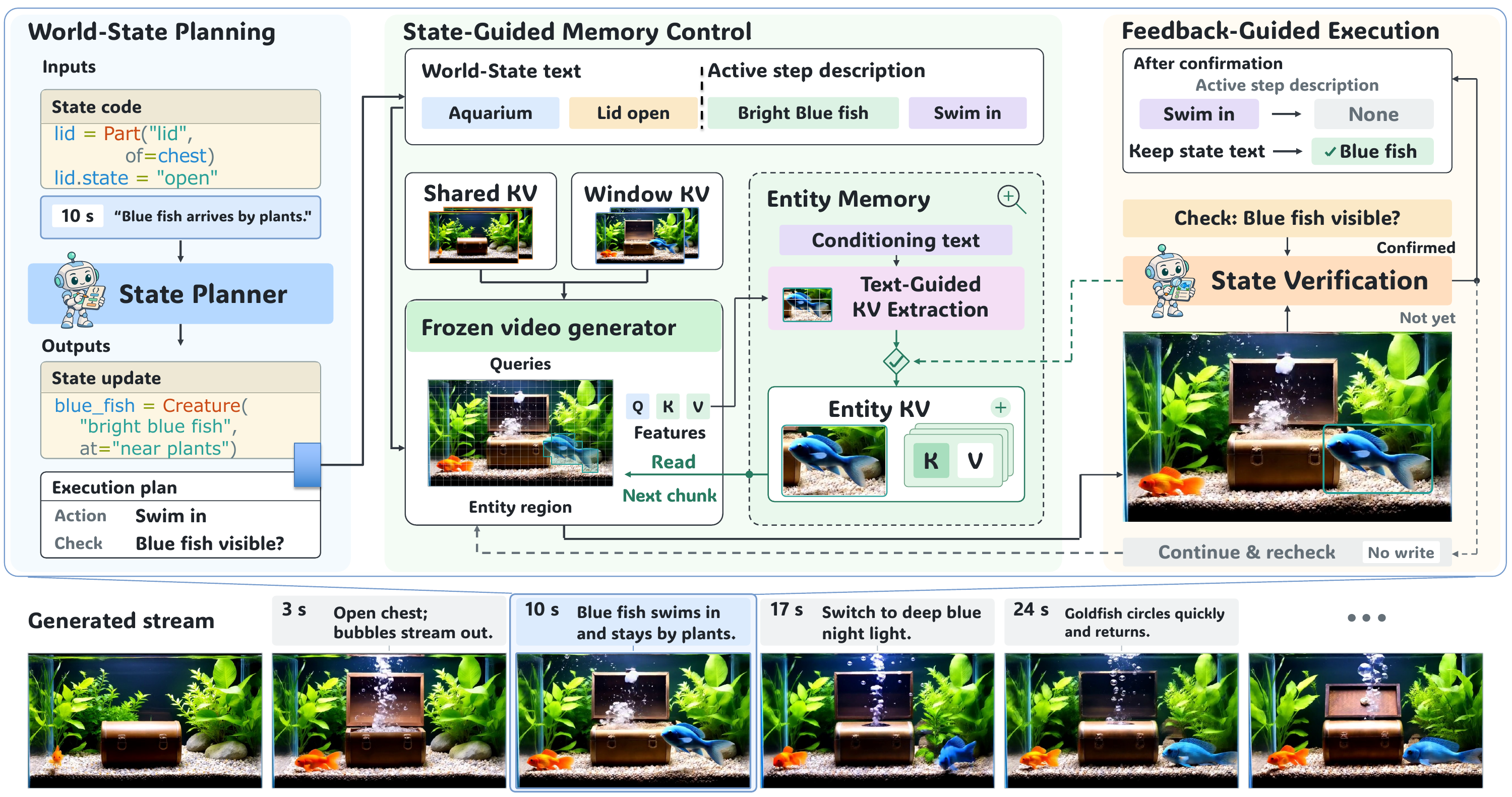}
    \caption{\textbf{Overview of \method{}.}
    \textbf{World-State Planning} uses the State Planner to turn state code and a new interaction into a state update and execution plan.
    \textbf{State-Guided Memory Control} combines World-State text and the active step description with Shared KV, Window KV, and Entity Memory.
    \textbf{Feedback-Guided Execution} checks generated frames to guide step progression and authorize local memory updates; completed interactions retain their state descriptions. Fig.~\ref{fig:entity_memory} details Entity Memory.}
    \label{fig:pipeline}
\end{figure*}

\section{Method}
\label{sec:method}
\label{sec:interface}

Given an initial description $d_0$ and interactive prompts $c_i$ arriving at times $a_i$, \method{} guides a frozen autoregressive generator $G_\theta$~\cite{longlive} to synthesize latent video chunks $x_t$ through three modules sharing a world-state representation (Fig.~\ref{fig:pipeline}).
\textbf{World-State Planning} interprets each prompt using the state program and produces requested state updates and an execution plan.
\textbf{State-Guided Memory Control} translates these updates into conditioning text and controls access to Entity KV.
\textbf{Feedback-Guided Execution} checks generated frames to govern step progression and enable local memory updates after confirmation.

\subsection{World-State Planning}
\label{sec:planning}
\label{sec:state}
\label{sec:reasoning}

\noindent\textbf{State representation and updates.}
We represent the initial description as a high-level state program $P_0$. Named variables represent entities and editable scene components, while \texttt{world} records global conditions such as lighting and weather. Given an interactive prompt $c_i$, our \emph{State Planner}, a VLM-based planning agent~\cite{qwen3vl}, interprets the request using the existing state program and produces a requested state update $\Delta_i$ and an execution plan $\pi_i$:
\begin{equation}
\begin{aligned}
(\Delta_i,\pi_i)&=\mathcal{A}(c_i,P_{i-1}),\\
P_i&=\operatorname{Append}(P_{i-1},\Delta_i),
\end{aligned}
\label{eq:reason}
\end{equation}
$\mathcal{A}$ denotes State Planner's interpretation and plan construction, $\operatorname{Append}$ adds the requested operations in $\Delta_i$ while preserving the program history. The resulting $P_i$ records requested transitions for interpreting subsequent interactions. 
Visual confirmation is kept in a separate execution state that governs step progression and memory updates (Sec.~\ref{sec:feedback}). 
This separation keeps the requested transition history independent of visual verification errors.

\noindent\textbf{State-dependent reasoning.}
The State Planner uses the state program and its transition history to resolve entity references in memory-dependent prompts. It reasons over recorded entity attributes and global conditions to determine the state updates required by the current interaction. External conditions requiring future visual observation become waiting steps, which Feedback-Guided Execution checks during generation.

\noindent\textbf{Execution plans.}
The execution plan $\pi_i$ organizes the interaction into ordered steps according to action dependencies. Each step specifies an action description, the affected entity variables, an intended result, a nominal duration, and a visual question with its expected answer. The descriptions and entity targets guide generation, while visual checks guide execution progress. Waiting steps omit the action description and check external conditions without prompting the generator to create them.

\subsection{State-Guided Memory Control}
\label{sec:generation}
\label{sec:rule}

\begin{figure}[t]
    \centering
    \includegraphics[width=\columnwidth]{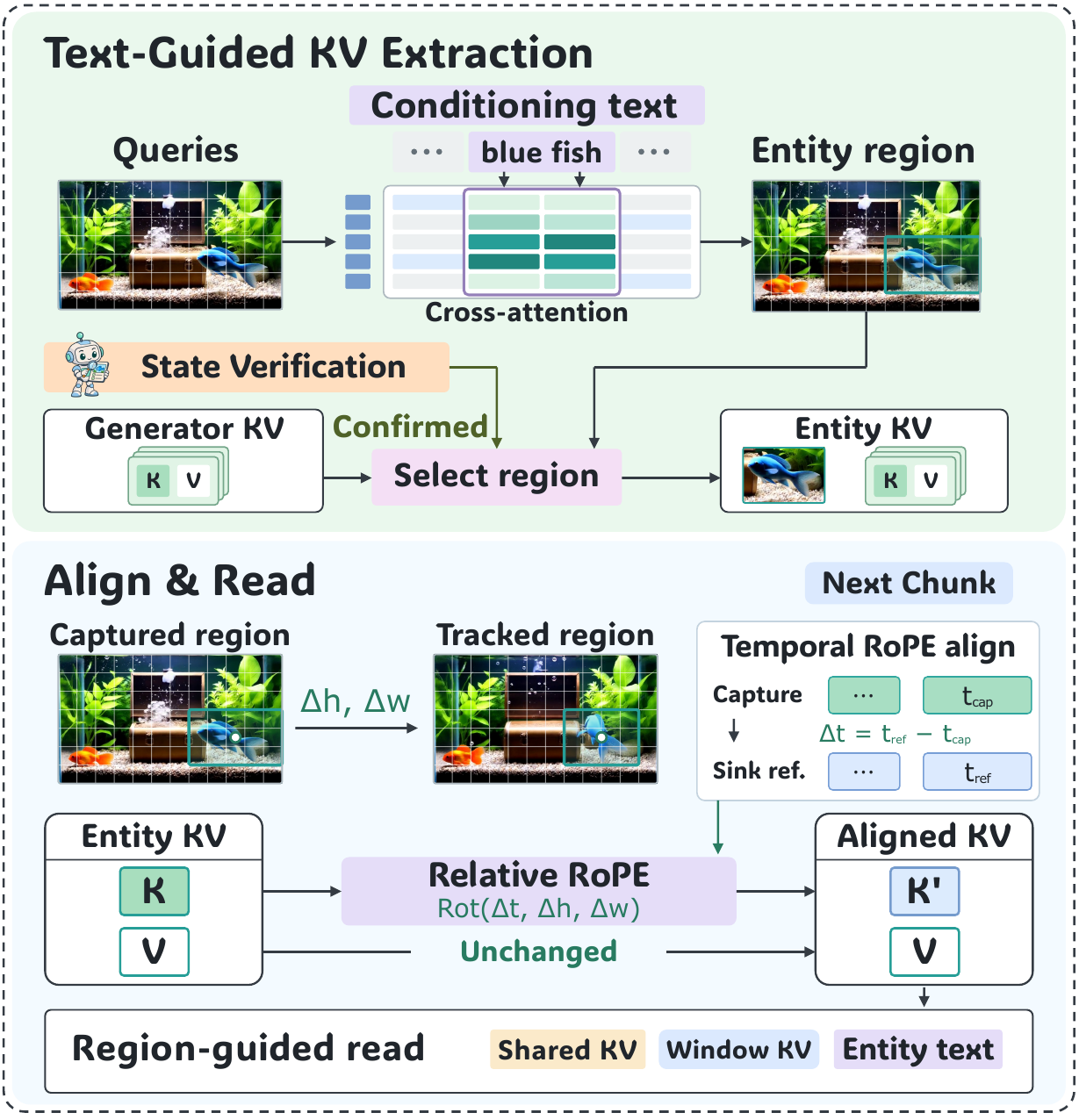}
    \caption{\textbf{Entity Memory.} Cross-attention localizes the entity region; visual confirmation authorizes Text-Guided KV Extraction. Relative RoPE shifts stored K to the sink reference time and tracked location, leaving V unchanged. Region-guided reading combines Entity KV with Shared / Window KV, and entity text.}
    \label{fig:entity_memory}
\end{figure}

State-Guided Memory Control translates requested state updates into conditioning text and controls the generator's access to Entity KV. \emph{Entity Memory} carries accepted local visual evidence into subsequent generation through \emph{Text-Guided KV Extraction} and \emph{Align \& Read} (Fig.~\ref{fig:entity_memory}).

\noindent\textbf{World-State text and active step descriptions.}
Conditioning text combines persistent scene and entity descriptions (\emph{World-State text}) with \emph{active step descriptions}. Using execution state $z_t$, which records execution plans, step progress, and confirmation, we construct:
\begin{equation}
y_t=\mathcal{T}(P_{i(t)},z_t),
\label{eq:overview}
\end{equation}
where $i(t)$ counts interactive prompts received before chunk $t$ and $\mathcal{T}$ assembles conditioning text. Target attributes may enter World-State text before visual confirmation to guide generation toward the requested state. After a final action's planned duration, its intended result description can replace the active step description. We retain this description during continued verification and after confirmation, unless a later interaction changes the target state. For actions requiring an external condition, the corresponding active step description is withheld until that condition is confirmed. Departure updates remove entities from World-State text.

\noindent\textbf{Text-Guided KV Extraction.}
Visual memory comprises \emph{Shared KV} $B_t$, which anchors the initial scene and permits local revision; \emph{Window KV} $W_t$ for recent dynamics; and \emph{Entity KV} $E_e$ for entity appearance. 
For initial entities that move or leave the scene, we temporarily copy their regional K/V from Shared KV into Entity KV entries for spatial alignment.
We use the generator's cross-attention maps from visual tokens to entity words in the conditioning text to localize an \emph{entity region}~\cite{prompt2prompt}. For local memory updates, we extract visual K/V at these positions from the generator's recent self-attention cache only after State Verification confirms the target result. The extracted K/V initialize or update Entity KV entries, or revise the corresponding Shared KV region for initial entities (Sec.~\ref{sec:feedback}).

\noindent\textbf{Align \& Read.}
For stored Entity KV $(K_e,V_e)$, the keys encode their capture coordinates. Before each chunk, we align them to the temporal reference range of Shared KV and the tracked entity region or active movement target. Given the offset $\Delta p_e=(\Delta t_e,\Delta h_e,\Delta w_e)$ from capture to reference coordinates, relative RoPE~\cite{roformer} gives
\begin{equation}
\widetilde K_e=\operatorname{Rot}(\Delta p_e)K_e,
\qquad \widetilde V_e=V_e.
\label{eq:entity_alignment}
\end{equation}
Here, $\operatorname{Rot}$ denotes the RoPE rotation. The shift preserves the temporal spacing of stored latent frames and leaves values unchanged.
We concatenate the aligned Entity KV with Shared KV and Window KV for the self-attention blocks. Binary masks $R_e(q)$ and $D_t(k)$ mark queries within entity $e$'s region and outdated Shared KV positions during movement or removal, respectively. 
With fixed positive bias strengths $\beta_s$, $\beta_e$, and $\gamma$, we add:
\begin{equation}
b_t(q,k)=
\begin{cases}
-\beta_sD_t(k), & k\in B_t,\\
\gamma R_e(q)-\beta_e(1-R_e(q)), & k\in E_e,\\
0, & k\in W_t.
\end{cases}
\label{eq:rule}
\end{equation}
to the attention logits. This enhances entity references within their regions and attenuates them elsewhere, while downweighting outdated Shared KV. Setting $R_e=0$ everywhere reduces the impact of an inactive entity reference across frames. 
Cross-attention similarly lowers the logits for bound entity words outside their regions. Global descriptions remain available throughout frames, and changes update World-State text while preserving Shared KV.


\begin{table*}[t]
\centering
\small
\setlength{\tabcolsep}{3.5pt}

\caption{\textbf{Interactive generation performance} across tasks and generators (\%).
LLM rewriter uses Qwen3-VL~\cite{qwen3vl} to rewrite interactive
prompts into action and scene descriptions for LongLive.
SF / LL denote Self-Forcing / LongLive; comparisons use matched
observations within each backbone.
Best and second-best results in each column are
\textbf{bold} and \underline{underlined}, respectively, including ties.
Superscripts show comparison with corresponding
baselines:
\textcolor{red}{improvement}, \textcolor{gray}{tie}, and
\textcolor{green!50!black}{degradation}.
Evaluation is defined in Sec.~\ref{sec:exp_setup}.}

\label{tab:main}
\resizebox{\linewidth}{!}{%
\begin{tabular}{@{}l*{9}{c}@{}}
\toprule
& \multicolumn{2}{c}{Basic}
& Dependent
& Conditional
& \multicolumn{2}{c}{Paired}
& New scenes
& \multicolumn{2}{c}{60\,s (ISR $\uparrow$)} \\
\cmidrule(lr){2-3}
\cmidrule(lr){6-7}
\cmidrule(lr){9-10}
Method
& ISR $\uparrow$
& TER $\uparrow$
& ISR $\uparrow$
& ISR $\uparrow$
& PER $\uparrow$
& PRR $\uparrow$
& ISR $\uparrow$
& Continuation
& Interaction \\
\midrule
TempAct~\cite{tempact}
& 26.72 & 37.02 & 13.95 & 12.62 & 12.50
& \underline{8.33} & 32.10 & 25.76 & 19.82 \\
IAMFlow~\cite{iamflow}
& 28.24 & 39.69 & 10.47 & 10.68 & 4.17
& 4.17 & 27.16 & 28.03 & 22.52 \\
MemFlow~\cite{memflow}
& 27.10 & 39.31 & 10.47 & 7.77 & 8.33
& 4.17 & 28.40 & 28.03 & 20.27 \\
SlotMemory~\cite{slotmemory}
& 28.63 & 33.59 & 15.12 & 14.56 & 8.33
& \underline{8.33} & 24.69 & 23.48 & 20.27 \\
LongLive-RAG~\cite{longliverag}
& 27.10 & 28.63 & 13.95 & 21.36 & 0.00
& 0.00 & 30.86 & 22.73 & 18.47 \\
LLM rewriter (LL)
& 36.64 & 37.02 & 19.77 & 33.01
& \underline{16.67}
& \underline{8.33}
& 33.33 & 37.88 & 39.19 \\
\midrule
Self-Forcing~\cite{selfforcing}
& 24.81 & 34.73 & 11.63 & 11.65 & 12.50
& \underline{8.33} & 27.16 & 28.79 & 22.07 \\
\method{} (SF)
& \underline{48.85}\metricgain{red}{+24.05}
& \underline{45.42}\metricgain{red}{+10.69}
& \textbf{37.21}\metricgain{red}{+25.58}
& \underline{38.83}\metricgain{red}{+27.18}
& \underline{16.67}\metricgain{red}{+4.17}
& \textbf{12.50}\metricgain{red}{+4.17}
& \underline{39.51}\metricgain{red}{+12.35}
& \underline{47.73}\metricgain{red}{+18.94}
& \underline{50.00}\metricgain{red}{+27.93} \\
\midrule
LongLive~\cite{longlive}
& 26.34 & 37.79 & 13.95 & 10.68 & 8.33
& 4.17 & 32.10 & 24.24 & 19.37 \\
\method{} (LL)
& \textbf{50.00}\metricgain{red}{+23.66}
& \textbf{48.09}\metricgain{red}{+10.31}
& \underline{34.88}\metricgain{red}{+20.93}
& \textbf{45.63}\metricgain{red}{+34.95}
& \textbf{20.83}\metricgain{red}{+12.50}
& \textbf{12.50}\metricgain{red}{+8.33}
& \textbf{46.91}\metricgain{red}{+14.81}
& \textbf{52.27}\metricgain{red}{+28.03}
& \textbf{50.45}\metricgain{red}{+31.08} \\
\bottomrule
\end{tabular}%
}
\end{table*}

\subsection{Feedback-Guided Execution}
\label{sec:feedback}
\label{sec:runtime}

\emph{State Verification} uses the same VLM to answer the execution plan's visual questions from generated frames. Its decisions are recorded in execution state $z_t$ and used to advance steps or enable local memory updates.

\noindent\textbf{Confirmation and step progression.}
We track each interaction's progress independently. Actions that depend on external conditions begin only after those conditions are visually confirmed. 
For interactions with lasting effects, we check each action step after its planned duration. 
Confirmation moves execution to the next step, or completes the interaction at the final step. If an intermediate action reaches its time limit without confirmation, execution proceeds to the next step, with its outcome recorded as \emph{unverified} in $z_t$. 
Waiting steps require confirmation and cannot be skipped
on timeout.
After an interaction completes, its result remains in World-State text, while the active step description is no longer supplied to the generator (Fig.~\ref{fig:pipeline}). For actions without lasting effects, we stop supplying the active step description after the planned duration.

\noindent\textbf{Local memory updates.}
For a local update, we replace only the selected Shared KV region with recent K/V after confirming the final result. With confirmation indicator $\eta_i\in\{0,1\}$ and binary region mask $M_i$ estimated from cross-attention, we write:
\begin{equation}
B^+=(1-\eta_iM_i)\odot B^-
       +\eta_iM_i\odot\widehat C_t,
\label{eq:write}
\end{equation}
where $B^-$ and $B^+$ are Shared KV before and after the update, $\widehat C_t$ contains K/V from the latest generated block with original RoPE encoding. The same mask applies to K and V across frames and layers. For separate entities, Text-Guided KV Extraction initializes or updates $E_e$ after confirmation.
After departure, we mark the entity's reference inactive or replace its Shared KV region with observed background. Actions without lasting effects create no persistent entries. Confirmation applies only to the condition checked.

\noindent\textbf{Continue \& recheck.}
Unconfirmed final results are rechecked against recent frames at their time limit. If still \emph{unverified}, their target descriptions are retained while subsequent chunks are checked. New persistent evidence is written only after confirmation; Window KV and other interactions continue updating.

\section{Experiments}
\label{sec:exp}

To measure the task of interactive long video generation, we propose a novel evaluation protocol that measures interaction execution and state retention.
We evaluate \method{} and other state-of-the-art methods on this novel protocol. Moreover, we evaluate the video quality using metrics in VBench. Quantitative comparisons, qualitative examples, and ablations examine how world-state reasoning and memory control support sustained interactions.

\begin{figure*}[t]
    \centering
    \includegraphics[width=\textwidth]{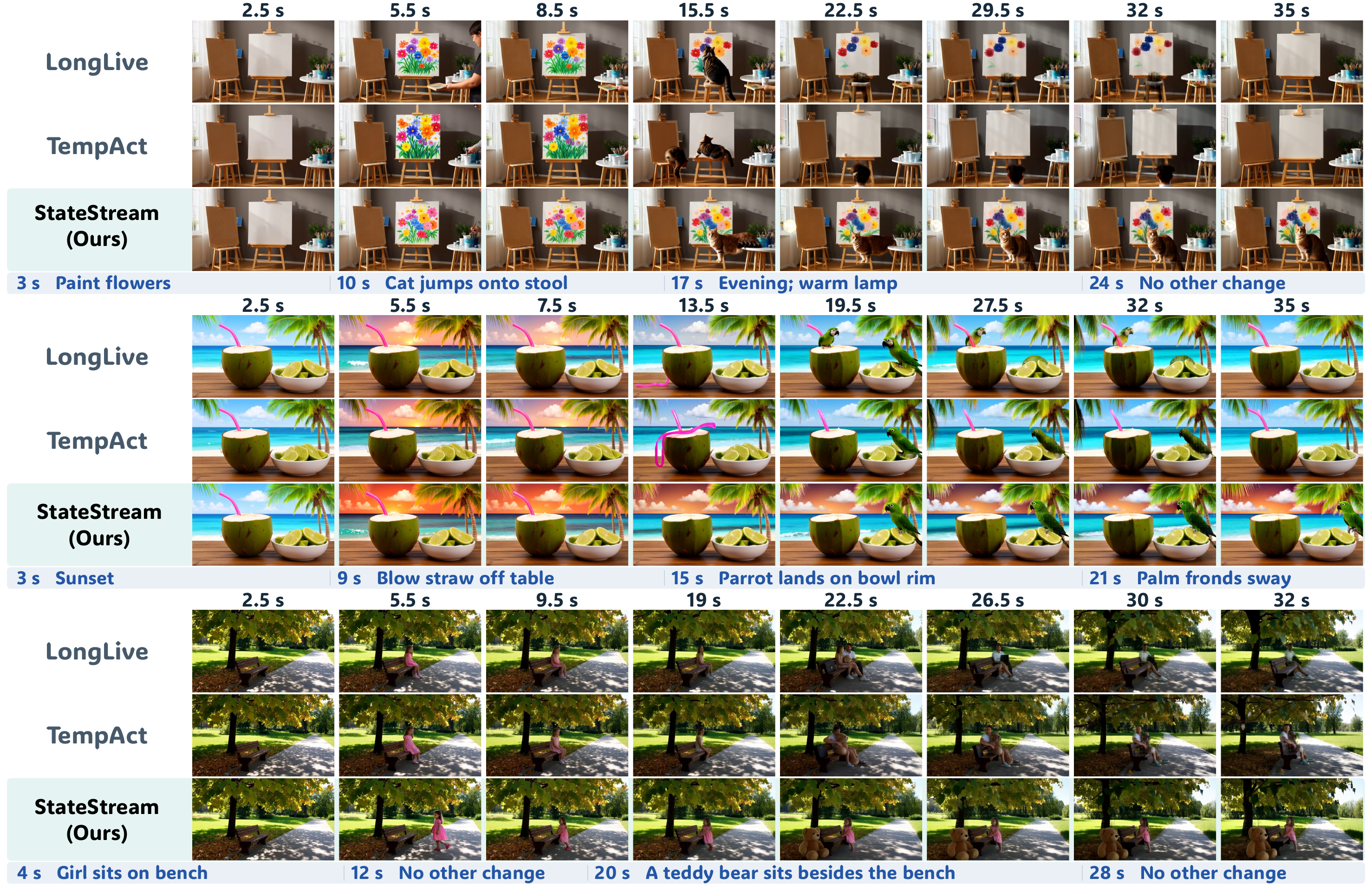}
    \caption{\textbf{Qualitative comparison} across successive interactions.
    Methods share initial scenes, interactive prompts, and sampling times. Later frames reveal differences in retaining the effects of earlier interactions.}
    \label{fig:qualitative_results}
\end{figure*}

\subsection{Experimental Setup}
\label{sec:exp_setup}

\noindent\textbf{Evaluation protocol.}
Each evaluation script specifies an initial scene description, interactive prompts with arrival times, and visual questions at predefined evaluation times.
We evaluate basic interactions (\emph{Basic}) in 70 scenes with four prompts per 36\,s video, covering local and global changes, entity addition and removal, and brief actions such as jumping.
Additional evaluations test action sequences with prerequisites and ordering constraints (\emph{Dependent}), actions conditioned on scene states or external triggers (\emph{Conditional}), and interactions in new scenes adapted from VBench prompts (\emph{New scenes}).
\emph{Paired} evaluation tests responses to identical target prompts under different histories from the same initial scene, including cases resolvable from the current frame.
Two 60\,s settings evaluate retention without further prompts (\emph{Continuation}) and with additional prompts (\emph{Interaction}) on different scenes.

\noindent\textbf{Evaluation metrics.}
\emph{Interaction Success Rate} (ISR) measures overall interaction correctness. Requested state changes must be observed at a predefined evaluation time and pass relevant entity-count checks. These changes must also hold at video end unless later prompts request otherwise.
\emph{Timely Execution Rate} (TER) checks whether a state change is visible 6\,s after its interactive prompt; ISR allows realization at any applicable check, including
the final check, without requiring all intermediate checks to pass.
\emph{Paired Execution Rate} (PER) is the fraction of pairs in which both responses pass execution checks; \emph{Paired Retention Rate} (PRR) additionally requires both outcomes to remain correct at the later evaluation.
We use the proprietary VLM~\cite{openai2026gpt6sol} to answer predefined questions on sampled frames. For visual quality, we average per-video scores for six VBench~\cite{vbench} dimensions: subject consistency (SC), background consistency (BC), motion smoothness (MS), dynamic degree (DD), aesthetic quality (AQ), and imaging quality (IQ).

\noindent\textbf{Implementation details.}
\method{} uses frozen LongLive-1.3B~\cite{longlive} and Self-Forcing~\cite{selfforcing} as video generators and Qwen3-VL-8B~\cite{qwen3vl} as the agent, without task-specific training. We generate $832\times480$ videos at 16\,fps. Full implementation and metric definition details are provided in Appendices~\ref{sec:supp_implementation} and~\ref{sec:supp_evaluation}.

\noindent\textbf{Baseline Methods.}
We compare with LongLive~\cite{longlive}, Self-Forcing~\cite{selfforcing}, TempAct~\cite{tempact}, IAMFlow~\cite{iamflow}, MemFlow~\cite{memflow}, SlotMemory~\cite{slotmemory}, LongLive-RAG~\cite{longliverag}, and a history-aware LLM rewriter. The rewriter uses the same language model and preceding interactions to update conditioning text, without visual feedback or additional memory control. All methods share initial descriptions, interactive prompts, and prompt arrival times.

\subsection{Quantitative Results}
\label{sec:exp_comparison}
\label{sec:exp_quantitative}
\noindent\textbf{Interaction evaluation.}
Table~\ref{tab:main} shows higher ISR for \method{} (LL) than LongLive, TempAct, IAMFlow, and MemFlow across different interactive scenarios. Gains over the LLM rewriter show that our method improves substantially over prompt engineering alone. With Self-Forcing, \method{} improves basic and long-horizon ISR, as well as paired execution or retention for the interactions in long video generation.

\begin{table}[t]
\centering
\footnotesize
\setlength{\tabcolsep}{2.2pt}

\caption{\textbf{Visual quality} on the basic interaction suite.
Best and second-best results are
\textbf{bold} and \underline{underlined}, respectively.}
\label{tab:vbench}

\begin{tabular*}{\linewidth}{@{\extracolsep{\fill}}l*{6}{c}@{}}
\toprule
Method & SC & BC & MS & DD & AQ & IQ \\
\midrule
TempAct
& 95.81 & 95.44 & 99.08 & 21.43 & 60.06 & 73.10 \\
IAMFlow
& 95.63 & \textbf{96.05} & 99.01 & 17.55
& \underline{61.16} & 71.95 \\
MemFlow
& 95.96 & 95.47 & 99.06 & 24.69
& 59.31 & \underline{73.25} \\
SlotMemory
& 93.13 & 93.80 & 98.21 & \textbf{37.76}
& 58.34 & 72.71 \\
LongLive-RAG
& 92.67 & 94.14 & 98.88 & 8.98 & 57.34 & 62.13 \\
LLM rewriter (LL)
& \underline{95.98} & 95.61 & \underline{99.10}
& 22.04 & 60.82 & 72.90 \\
\midrule
Self-Forcing
& 95.32 & 95.01 & 98.87 & 26.33 & 60.01 & 71.95 \\
\method{} (SF)
& 95.45 & 95.13 & 98.91 & 26.53 & 59.96 & 72.82 \\
\midrule
LongLive
& 95.81 & 95.44 & 99.08 & 21.02 & 60.61 & 73.14 \\
\method{} (LL)
& \textbf{96.09}
& \underline{95.90}
& \textbf{99.14}
& \underline{30.41}
& \textbf{61.52}
& \textbf{74.46} \\
\bottomrule
\end{tabular*}
\end{table}

\noindent\textbf{Visual quality.}
Table~\ref{tab:vbench} shows that VBench scores remain comparable to or better than the state-of-the-art methods, especially in consistency, motion smoothness, aesthetics, and image quality due to the more natural interaction control. Note that SlotMemory shows a higher dynamic degree than ours; however, we observe unnecessary camera movements and frame drifting in many of its generated videos.

\begin{table}[!tb]
\centering\small
\setlength{\tabcolsep}{4pt}
\caption{User study of generated long videos and interactions.}
\label{tab:user}
\begin{tabular*}{\columnwidth}{@{\extracolsep{\fill}}lcccc@{}}
\toprule
Method & Exec. $\uparrow$ & Ret. $\uparrow$ & VQ $\uparrow$ & Coh. $\uparrow$ \\
\midrule
TempAct & 3.68 & 3.42 & 3.36 & 3.54 \\
IAMFlow & 3.20 & 3.08 & 3.76 & 3.62 \\
LongLive-RAG & 3.10 & 2.96 & 3.60 & 3.06 \\
LLM rewriter & 3.34 & 3.58 & 3.74 & 3.24 \\
\midrule
LongLive & 3.12 & 3.08 & 3.72 & 3.04 \\
\method{} (LL) & \textbf{3.84} & \textbf{4.06} & \textbf{3.86} & \textbf{3.66} \\
\bottomrule
\end{tabular*}
\end{table}

\noindent\textbf{User study.}
We also conduct a user study to provide complementary human evidence to our evaluation. We evaluate: Execution, Retention, Visual Quality, and Interaction Coherence of generated videos. Each aspect is rated from 1 to 5 (higher is better). As shown in Tab.~\ref{tab:user}, the user ratings align with our evaluation protocol, showcasing our \method{}'s capability in generating interactive long videos. See Appendix~\ref{sec:supp_evaluation} for more details.

\begin{figure*}[t]
    \centering
    \includegraphics[width=\textwidth]{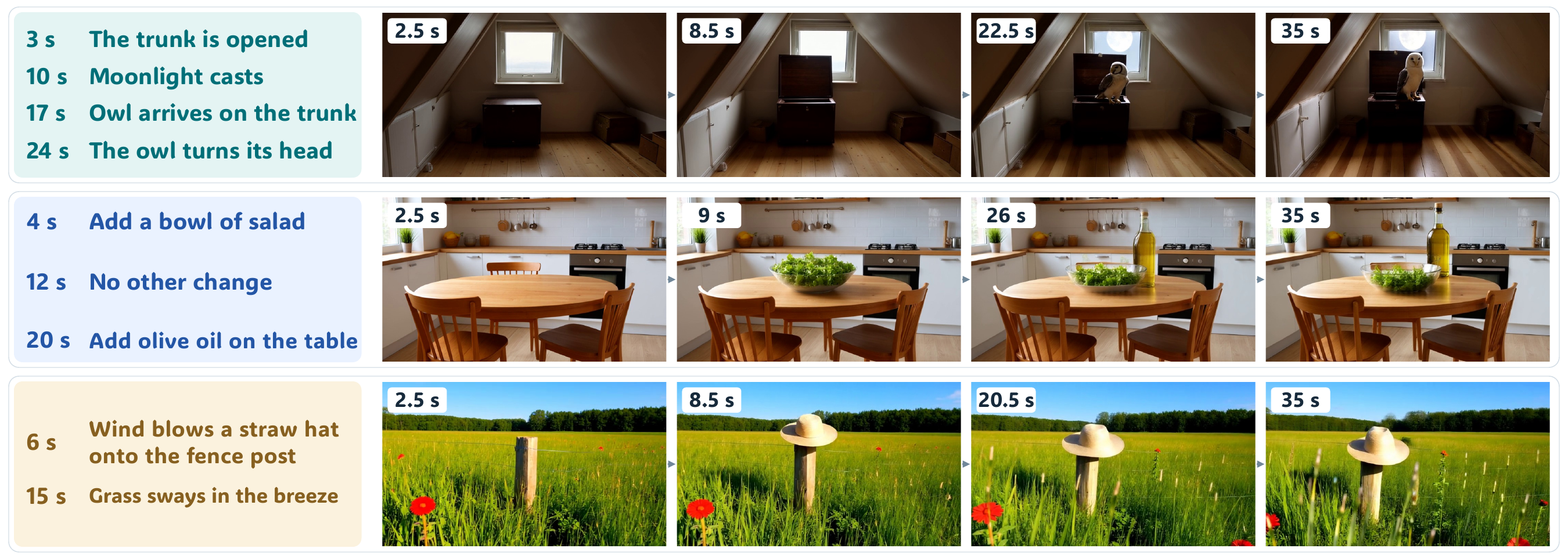}
    \caption{\textbf{Sequential interaction with evolving world state.}
    Representative sequences show how new interactions build on earlier outcomes during continued generation. Prompts and sampling times indicate the progression of each video.}
    \label{fig:interactive_gallery}
\end{figure*}

\subsection{Qualitative Results}
\label{sec:exp_qualitative}

\begin{figure}[t]
    \centering
    \includegraphics[width=\columnwidth]{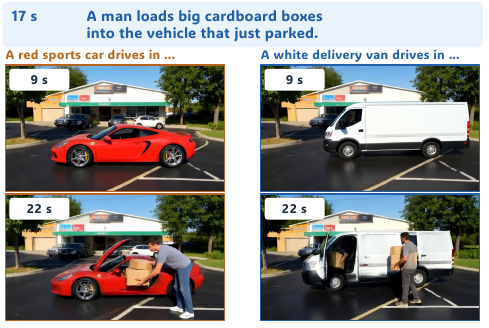}
    \caption{Different interactions result in different outcomes.}
    \label{fig:history_response}
\end{figure}

In Fig.~\ref{fig:qualitative_results}, \method{} more consistently carries realized changes into later frames, whereas the compared baselines can lose previously generated content as generation continues. Figure~\ref{fig:interactive_gallery} further illustrates how new interactions build on the evolving scene while retaining earlier outcomes that remain applicable. These results highlight the role of world state in guiding generation.
Figure~\ref{fig:history_response} further illustrates state-dependent target selection, with the same interactive prompt referring to different entities under different interaction histories. Together, these examples highlight how shared world state helps both interpreting new interactions in context and retaining their consequences during continued generation.

\subsection{Ablation Study}
\label{sec:exp_ablation}

\begin{table}[t]
\centering\footnotesize
\setlength{\tabcolsep}{3pt}
\caption{Ablation study.
ISR, TER, and DR are evaluated on the basic suite; Cond. ISR on conditional events.}
\label{tab:ablation}
\begin{tabular}{@{}lcccc@{}}
\toprule
Variant & ISR $\uparrow$ & TER $\uparrow$
& Dup. Rate $\downarrow$ & Cond. ISR $\uparrow$ \\
\midrule
Ours Full
& 50.00 & 48.09 & 13.33 & 45.63 \\
$-$State-Guided Memory
& 43.89 & 45.42 & 24.32 & 42.72 \\
$-$Feedback-Guided Exec.
& 49.62 & 44.27 & 18.42 & 40.78 \\
$-$Execution Plans
& 41.22 & 46.95 & 13.89 & 30.10 \\
\midrule
w/o Local Memory Updates
& 45.42 & 42.37 & 12.50 & 43.69 \\
w/o Write Confirmation
& 46.56 & 45.80 & 15.58 & 43.69 \\
\bottomrule
\end{tabular}
\end{table}

\begin{figure}[t]
    \centering
    \includegraphics[width=\columnwidth]{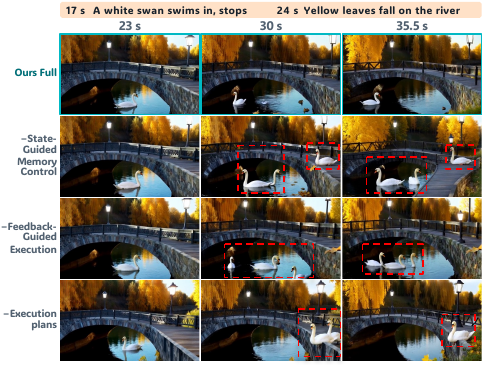}
    \caption{Visualization of ablation study across successive interactions. Red dashed boxes highlight artifacts in frames.}
    \label{fig:ablation_qualitative}
\end{figure}

Table~\ref{tab:ablation} and Fig.~\ref{fig:ablation_qualitative}
show complementary contributions from memory control, planning,
and feedback. DR measures how often a realized single arrival
later appears as multiple instances; conditional ISR evaluates
effects that must follow a triggering condition.
Removing State-Guided Memory Control reduces interaction success
and increases duplication. Removing Execution Plans causes the
largest drops in overall and conditional ISR, while removing
Feedback-Guided Execution lowers TER and conditional ISR,
supporting the roles of planning and feedback in execution timing.
Removing local memory updates reduces ISR and TER despite a
slightly lower DR, whereas removing write confirmation reduces
ISR and increases duplication.
Appendix~\ref{sec:supp_mechanism} provides further ablation details.

\subsection{Limitation}
\label{subsec:limitation}
Prompt adherence and visual quality depend on the frozen backbone's ability to realize requested changes. \method{} focuses on world-state reasoning, execution, and retention; improving the backbone's intrinsic prompt alignment is outside our scope. The modular framework could be adapted to stronger generators with compatible attention and KV-cache interfaces. Integration with commercial models remains future work.

\section{Conclusion}
We presented \method{}, an agentic framework for world-state reasoning and control in interactive long video generation.
On a frozen generator, a shared world-state representation connects memory-dependent prompt interpretation, execution planning, and persistent visual memory.
Visual feedback guides execution and local memory updates, distinguishing requested transitions from confirmed outcomes.
Together, these components connect interactions to persistent changes in the generated video.
Our results show improved interaction success and state retention, highlighting the value of state control, and marking a step toward generative worlds that respond to user interactions while preserving their effects over time.

{
    \small
    \bibliographystyle{ieeenat_fullname}
    \bibliography{main}
}

\clearpage
\pagenumbering{arabic}
\setcounter{topnumber}{2}
\maketitlesupplementary
\appendix

This supplementary material provides implementation and evaluation details, additional video examples, and agent prompt excerpts.

\section{Implementation Details}
\label{sec:supp_implementation}
Table~\ref{tab:supp_config} summarizes the settings shared across evaluation sets and backbones. The following details complement the three modules in Sec.~\ref{sec:method}.

\begin{table}[!b]
\caption{Core implementation settings.}
\label{tab:supp_config}
\centering\footnotesize
\setlength{\tabcolsep}{3pt}
\begin{tabular}{@{}lr@{}}
\toprule
Setting & Value \\
\midrule
\multicolumn{2}{@{}l}{\emph{Generator and streaming}} \\
Block length & 3 latent frames (0.75\,s) \\
Shared / Window / Entity KV & 3 / 12 / 3 latent frames \\
\midrule
\multicolumn{2}{@{}l}{\emph{Agent and Feedback-Guided Execution}} \\
Confirmation threshold & 0.7 \\
Consecutive confirmations $K$ & 2 \\
Check interval & 1 block \\
Planned step duration & 0.75--8\,s \\
\midrule
\multicolumn{2}{@{}l}{\emph{State-Guided Memory Control}} \\
Attention layer indices (zero-based) & 7, 8, 13, 18, 28 \\
Thresholds (interaction / extract / track) & 0.5 / 0.3 / 0.5 \\
Dilation (interaction / extract / read) & 1 / 1 / 2 tokens \\
Tracking radius / min. region size & 3 / 4 tokens \\
$\beta_s$, $\beta_e$, $\gamma$ & 5, 5, 1 \\
Entity-text penalty outside its region & 5 \\
\bottomrule
\end{tabular}
\end{table}

\subsection{Models and streaming configuration}
\label{sec:supp_models}
LongLive-1.3B~\cite{longlive} uses its released LoRA merged into frozen Wan2.1-T2V-1.3B weights. We generate $832\times480$ videos at 16\,fps using four denoising steps ($t=1000,750,500,250$; shift 5). An additional forward pass over the denoised latent frames computes K/V for memory updates. The nominal 36\,s and 60\,s settings produce 573 and 957 frames. When conditioning text changes, we recompute Window KV while keeping stored Shared KV and Entity KV features.
Qwen3-VL-8B-Instruct~\cite{qwen3vl} uses greedy decoding. A streaming VAE decodes one frame per block for State Verification. Generation runs on an RTX~6000 Ada (48\,GB), with the agent on a separate GPU for efficiency; a single-GPU setup is also supported. Execution times are reported in Sec.~\ref{sec:supp_cost}.

\noindent\textbf{Second backbone.}
\label{sec:supp_selfforcing}
Self-Forcing~\cite{selfforcing} uses the same Wan2.1 backbone. The baseline and \method{} use the same streaming settings, without retuning our method for this backbone. Evaluation includes an opening-frame sink reference, although Self-Forcing was trained without one.

\subsection{World-State Planning}
\label{sec:supp_planning}
Before video generation, we prepare programs and plans in prompt order, using only the current and preceding prompts. The state program records requested changes. During generation, visual checks update execution state and memory; they do not revise the program or produce new plans.
State code creates, modifies, and removes entities through constructors, assignments, and deletion. It uses \texttt{act} for temporary actions and \texttt{verify} for visual questions. These operations specify what changes, which entities are affected, and what to check.
Plans contain at most three steps. Required earlier actions precede the actions that depend on them. Steps waiting for external conditions provide no action description. Consecutive steps affecting the same entities are merged.

\subsection{State-Guided Memory Control}
\noindent\textbf{Conditioning text.}
\label{sec:supp_text}
World-State text is updated when a persistent change begins, after any required external condition is confirmed. We revise only the affected descriptions; they can express a target state before it is visually confirmed. Text checks correct missing updates and conflicting descriptions, skipping entities whose changes are still in progress.
After the planned duration, descriptions of persistent actions are replaced by descriptions of their intended results. Departure descriptions remain until confirmation or the time limit. Waiting for an external condition adds no action description, and temporary action descriptions end after their planned duration. After confirmation, cross-attention to entity words is reduced outside their regions. Descriptions of removed entities are deleted.

\noindent\textbf{Text-Guided KV Extraction.}
\label{sec:supp_regions}
We locate entities using cross-attention to their words, averaged over heads and selected layers. The resulting map guides attention from the next denoising step. To locate affected Shared KV tokens, we take the maximum at each spatial position over blocks generated during the interaction and normalize the map to $[0,1]$. To select tokens for Entity KV, we use the latest latent frame's attention map, subtract by its mean, set the negative values to zero, and divide by its maximum. We threshold and enlarge these regions using the hyper-parameters specified in Table~\ref{tab:supp_config}. If fewer than four extraction tokens remain, we use the region estimated from attention accumulated during the interaction.
For tracking, we average attention within each block and use the same mean-subtraction normalization near the previous region. Memory updates exclude other entities' regions if at least four extraction tokens remain. After confirmation, we extract K/V from the selected region in the latest three latent frames at every layer. Each entry stores these features, the entity ID, capture coordinates, and a mask marking the selected spatial tokens.

\noindent\textbf{Align \& Read.}
\label{sec:supp_access}
For reading, we shift the tracked region and apply it to every latent frame in the block. Appending dimensions to queries and keys implements the attention biases in Eq.~(\ref{eq:rule}) while keeping the original attention scale. Inactive entries stay in memory but receive less attention. For initial entities that move or leave, we create temporary Entity KV entries and reduce attention to their original Shared KV regions. Those regions are updated after confirmation.
We track motion from the center of the attention region. For speed $v\geq0.7$ tokens per block, the reference region is placed $\max(3,1.5v)$ tokens ahead in the motion direction. For departures, it moves at 5 tokens per block, at most $\max(4,0.7\sqrt{A})$ tokens ahead of the observed entity, where $A$ is its region size in tokens. If motion has not started after two blocks, the reference moves toward the nearer image side. Window KV tokens at vacated positions receive a $\ln 10^{-3}$ attention bias only in frames generated before departure, until those frames leave the window.
Relative RoPE aligns stored keys with the reference positions used for reading. A temporal shift of $2-t_{\rm cap}$ maps the three capture indices $(t_{\rm cap}-2,t_{\rm cap}-1,t_{\rm cap})$ to sink indices $(0,1,2)$. Spatial shifts move the reference from its captured region to its tracked region; values remain unchanged. Keys written locally to Shared KV keep their original temporal encoding. At latent indices 132 and 201, an additional $-\beta_s$ bias on Shared KV and temporally aligned Entity KV reduces the tendency to return to opening content~\cite{lol}.

\subsection{Feedback-Guided Execution}
\label{sec:supp_execution}
\label{sec:supp_lifecycle}
Each interaction progresses independently, one block at a time. A visual check passes when the verifier's confidence in the expected yes/no answer satisfies
\begin{equation}
p=\sigma(\ell_{\rm expected}-\ell_{\rm other})\geq0.7.
\label{eq:supp_yesno}
\end{equation}
Here, $\ell$ is computed from the first-token log probabilities of each answer. For multiple questions, we use the lowest confidence, so all questions must pass. Table~\ref{tab:supp_progress} specifies when steps finish or continue.
We check actions after their planned duration and external conditions from one block after the waiting step begins. A final step is the last action of an interaction with persistent results.
\begin{table}[t]
\caption{Step progression. $\tau$: planned duration; $K=2$. Time limits start at step onset.}
\label{tab:supp_progress}
\centering\footnotesize
\setlength{\tabcolsep}{2.5pt}
\begin{tabular}{@{}p{0.2\columnwidth}p{0.22\columnwidth}p{0.5\columnwidth}@{}}
\toprule
\raggedright Step & \raggedright Passing checks & \raggedright Without confirmation\tabularnewline
\midrule
\raggedright External condition & \raggedright $K$ in a row & \raggedright Keeps waiting; no timeout\tabularnewline
\raggedright Intermediate action & \raggedright 1 & \raggedright Starts next step at $\max(2\tau,3\,\mathrm{s})$; result stays unconfirmed\tabularnewline
\raggedright Final, arrival & \raggedright 1 & \raggedright Rechecks recent frames at 8.25\,s; unverified if checks fail\tabularnewline
\raggedright Final, other & \raggedright $K$ in a row & \raggedright Rechecks recent frames at 8.25\,s; unverified if checks fail\tabularnewline
\raggedright Transient & \raggedright -- & \raggedright Ends after its planned duration\tabularnewline
\bottomrule
\end{tabular}
\end{table}
Before an interaction, we test its questions on the current frame and revise or remove questions already satisfied. A presence question only checks whether an entity is visible; it does not verify all attributes or the number of instances.
If a final step reaches its time limit, we recheck frames from the last three blocks. Two consecutive frames must pass both the state program's and the step's questions. Otherwise, the result remains \emph{unverified}: we keep its target description and continue checking under the existing plan, without writing its K/V to persistent memory. Confirmed local results update Shared KV or Entity KV. Global changes keep Shared KV unchanged, and temporary actions create no persistent entries.

\subsection{Computation and memory}
\label{sec:supp_cost}
\label{sec:supp_runtime}
An entity region containing $m_e$ spatial tokens stores three latent frames, adding $3m_e$ K/V tokens per layer. Inactive entries stay in memory, so storage grows as interactions add entries. No separate image encoder is used.
Table~\ref{tab:supp_efficiency} reports execution time on the same 70 Basic scenes, including visual checks, text updates, and video decoding.

\begin{table}[t]
\caption{Execution time. Mean per video over 70 Basic scenes.}
\label{tab:supp_efficiency}
\centering\small
\setlength{\tabcolsep}{8pt}
\begin{tabular}{@{}lc@{}}
\toprule
Method & Execution time (s) \\
\midrule
LongLive & 141.6 \\
LLM rewriter & 141.8 \\
\method{} (LL) & 168.9 \\
\bottomrule
\end{tabular}
\end{table}

\begin{figure*}[h!]
\centering
\includegraphics[width=\textwidth]{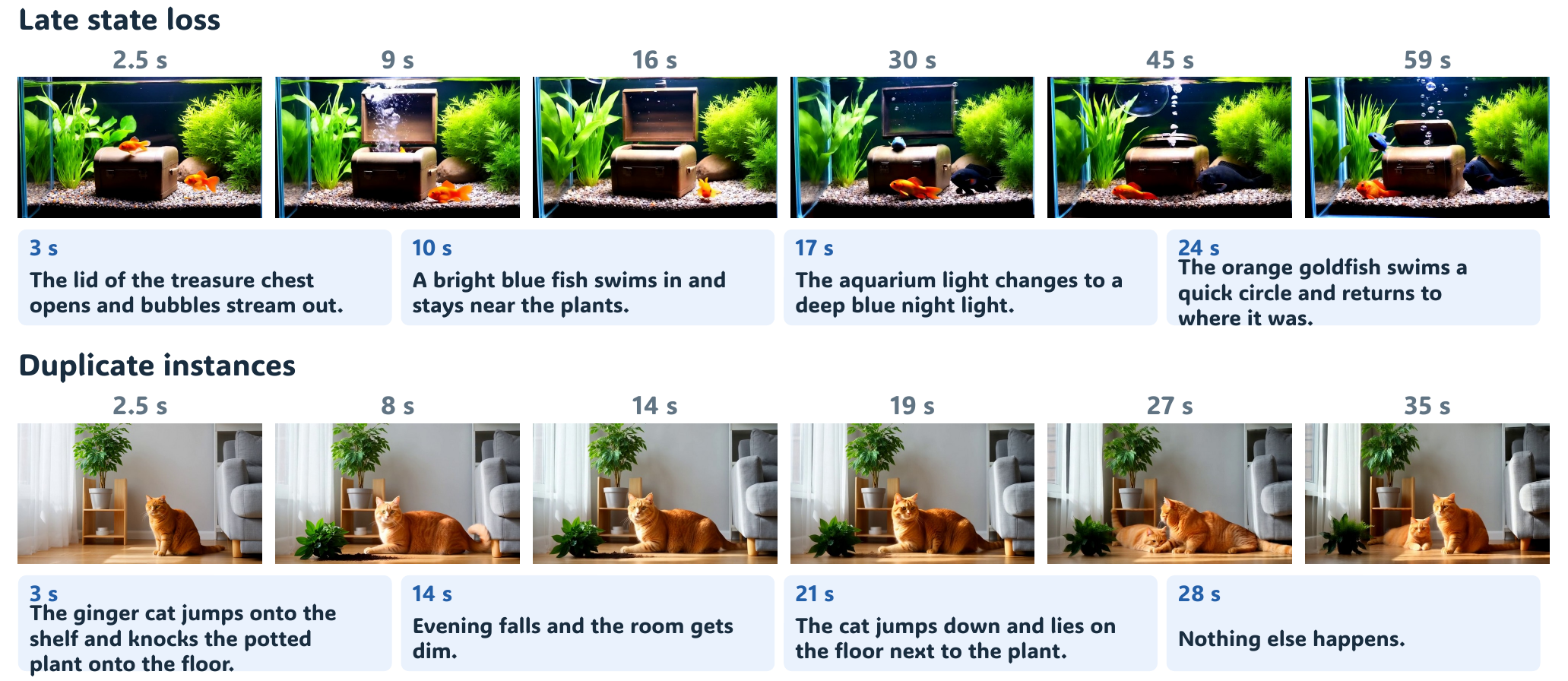}

\caption{Remaining failures. Top: state reversion and appearance drift. Bottom: entity duplication.}
\label{fig:supp_failures}

\end{figure*}

\section{Evaluation Details \& More Results}
\label{sec:supp_evaluation}
\subsection{Evaluation sets}
\label{sec:supp_suites}
Basic contains 70 scenes; Dependent, Conditional, and New scenes each contain 30 scripts; Paired contains 24 pairs. Paired scripts share an initial scene and target prompt but have different preceding interactions. They test whether responses adapt to the resulting state, including cases that can be resolved from the current frame.
Each 60\,s setting contains 35 scripts: Continuation has no new prompts after 24\,s, while Interaction adds prompts at 31, 38, and 45\,s. These settings use different scene subsets. 
The Basic suite contains 280 scheduled interactive prompts across 70 videos. Of these, 262 are included in the ISR and TER evaluation. The remaining 18 comprise six control prompts without task-specific success checks and 12 prompts whose required initial states fail the opening-state check on the LongLive reference videos. The same 262 prompts are used for all compared methods.

\subsection{Outcome metrics and judges}
We use the metrics in Sec.~\ref{sec:exp_setup}, where $a_i$ is the arrival time of an interactive prompt. 
%
For persistent effects, TER checks whether the requested change is visible at \(a_i+6\) s. ISR allows the effect to be first observed at any applicable evaluation time: \(a_i+6\) s, 20 s, 28 s, or the final check. If the effect remains applicable through the end of the video, it must also pass the final retention check. Retention checks at or after a later prompt that supersedes the effect no longer apply to the original request. For single-entity arrivals, ISR additionally requires that at most one instance of the entity be visible at the final check, even if a later prompt requests its departure. The departure itself is evaluated separately. Temporary actions must occur at \(a_i + 1.5\) s and return to rest at \(a_i + 5\) s.

\noindent For \textbf{Basic}, GPT 6-sol judges the evaluation frame and its $\pm0.5$\,s neighbors by majority vote, independently of the verifier used during generation. Dependent checks assess completion at $a_i+9$\,s and whether required earlier actions precede their effects. Conditional checks examine triggers and effects every 0.75\,s, require passing checks in consecutive frames, and allow one sampling interval of timing tolerance. Effects should remain absent when external conditions never occur. However, failing to generate a requested prerequisite does not count as success.

\noindent For \textbf{Paired}, both videos must show the required preceding state and correct response, without an incorrect alternative, at 26\,s (PER) and remain correct at 35.5\,s (PRR). DR checks for duplicate instances from $a_i+6$\,s onward among requests for one new entity that has appeared. VBench scores complete, non-overlapping 5\,s clips, averaging first within each video and then equally across videos.

\noindent\textbf{User study protocol.}
\label{sec:supp_judge_audit}
10 volunteers are asked evaluate the interactive long videos generated by different methods. Each participant evaluates 20 samples, with each sample presenting videos from all compared methods simultaneously in the same interactive prompts, with method names hidden and presentation order randomized. For each method, ratings are averaged across participants and samples to obtain the final score for each evaluation dimension.

\subsection{Ablation settings}
\label{sec:supp_ablation}
\emph{$-$Execution Plans} uses one action description per
interaction, retaining state updates, external-condition checks,
final-result verification, and confirmation-gated memory writes.
\emph{$-$State-Guided Memory Control} uses the backbone's
original KV cache while retaining state-conditioned text,
execution plans, and visual feedback.
\emph{$-$Feedback-Guided Execution} removes visual checks
and external-condition waits. Actions and memory writes follow
planned times; state updates, plans, and memory operations remain.
\emph{w/o Local Memory Updates} disables writes of new K/V
to Shared KV and Entity KV, retaining planning, verification,
and reference reading.
\emph{w/o Write Confirmation} performs each eligible local
write once at the final action's planned end without requiring
confirmation, retaining execution checks and reference reading.
\emph{World-State Planning only} removes visual memory
control and feedback, retaining state updates and plan-derived
text with execution at planned times.
All controls share the full method's backbone, scripts,
seeds, and evaluation protocol.

\subsection{Additional Ablations}
\label{sec:supp_mechanism}
Table~\ref{tab:supp_mechanism} reports two additional ablations. Temporal Alignment Only isolates the effect of spatial alignment in Entity Memory. World-State Planning Only retains state updates and plan-derived conditioning while removing visual memory control and feedback, as described in Sec.~\ref{sec:supp_ablation}.

\emph{Entity KV alignment.} In \emph{Temporal alignment only}, Entity KV is re-indexed in time but not shifted in space, so each entity reference remains at the position where it was captured rather than following the entity as it moves. This leads to noticeably more duplicated entities, while interaction success and TER decrease slightly. Spatial alignment is therefore important for keeping a single, consistent copy of each entity.

Compared with the full model, World-State Planning Only reduces ISR from 50.00\% to 44.27\% and increases the duplication rate from 13.33\% to 26.36\%, showing the combined contribution of memory control and feedback.

\begin{table}[t]
\caption{Additional ablations on the Basic suite (\%). All rows use the same evaluation protocol and Basic interactions as the main results.}
\label{tab:supp_mechanism}
\centering\footnotesize
\setlength{\tabcolsep}{3pt}
\begin{tabular*}{\columnwidth}{@{\extracolsep{\fill}}lcccc@{}}
\toprule
Setting & ISR $\uparrow$ & TER $\uparrow$ & DR $\downarrow$ & IQ $\uparrow$ \\
\midrule
\method{} (Full) & 50.00 & 48.09 & 13.33 & 74.46 \\
Temporal alignment only & 49.24 & 47.33 & 17.33 & 73.62 \\
World-State planning only & 44.27 & 46.56  & 26.36 & 72.58 \\
\bottomrule         
\end{tabular*}
\end{table}

\subsection{Additional Qualitative Results}
\label{sec:supp_qualitative}
\label{sec:exp_long}
\begin{figure*}[p]
\centering
\includegraphics[width=\textwidth]{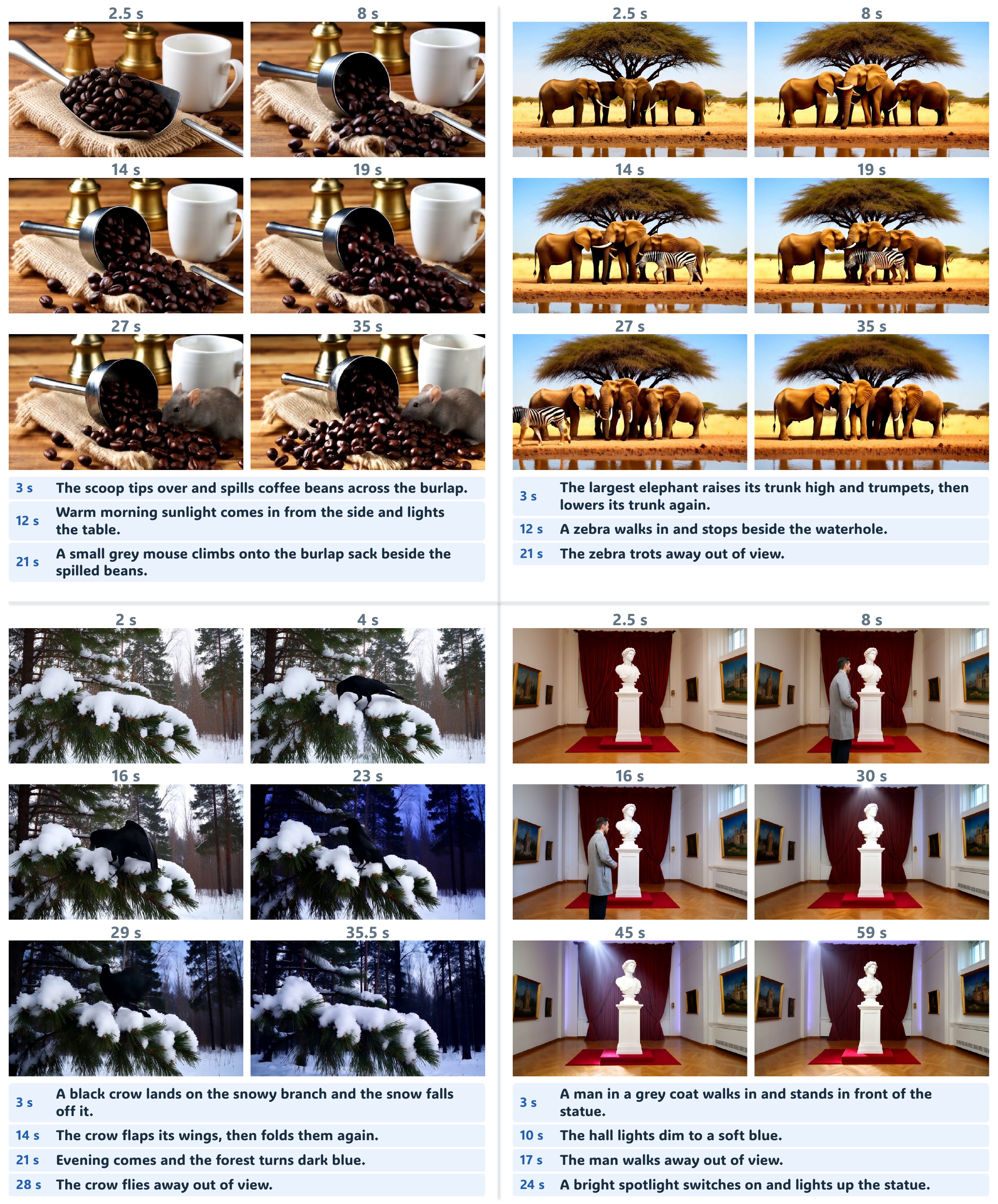}
\caption{Additional interactive results. Each panel shows one video, read left to right and then top to bottom.}
\label{fig:supp_qualitative}
\end{figure*}

\begin{figure*}[p]
\centering
\includegraphics[width=\textwidth,height=.905\textheight,keepaspectratio]{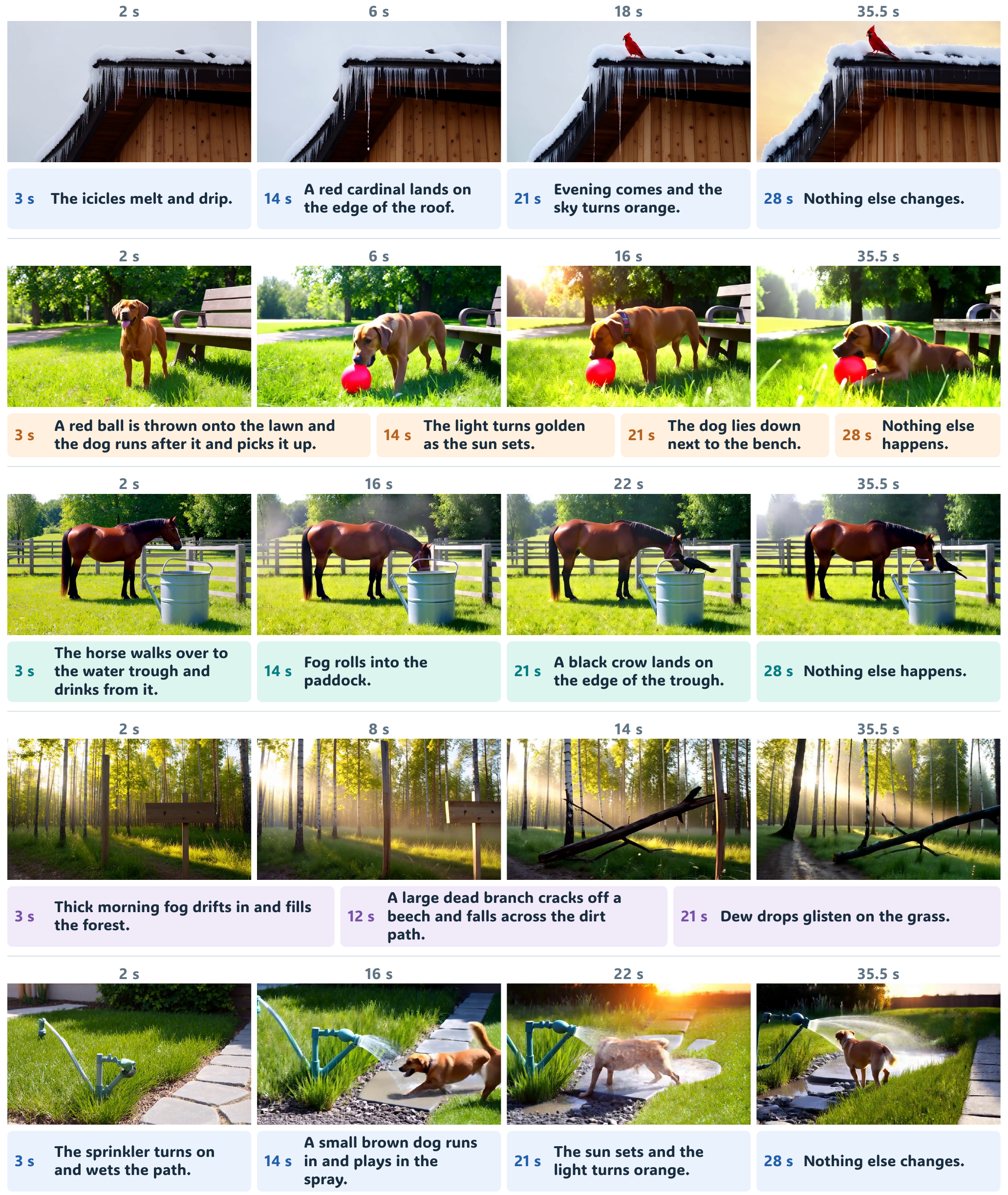}
\caption{Sequential interactions across scenes. Each row shows one video with its interactive prompts.}
\label{fig:supp_gallery_interactions}
\end{figure*}

\begin{figure*}[p]
\centering
\includegraphics[width=\textwidth,height=.905\textheight,keepaspectratio]{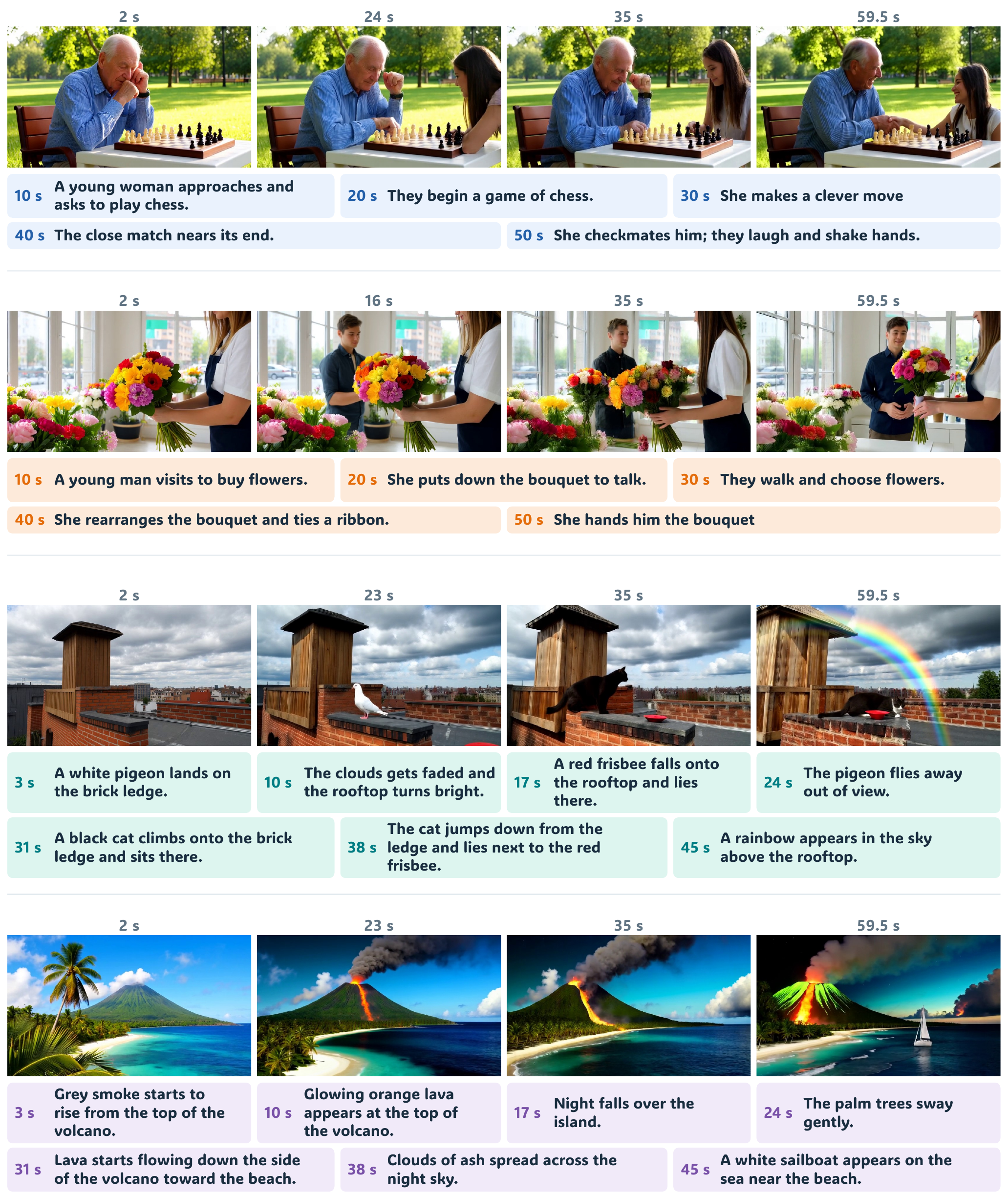}
\caption{Interactive generation over 60 seconds.}
\label{fig:supp_gallery_extended}
\end{figure*}

\begin{figure*}[p]
\centering
\includegraphics[width=\textwidth,height=.93\textheight,keepaspectratio]{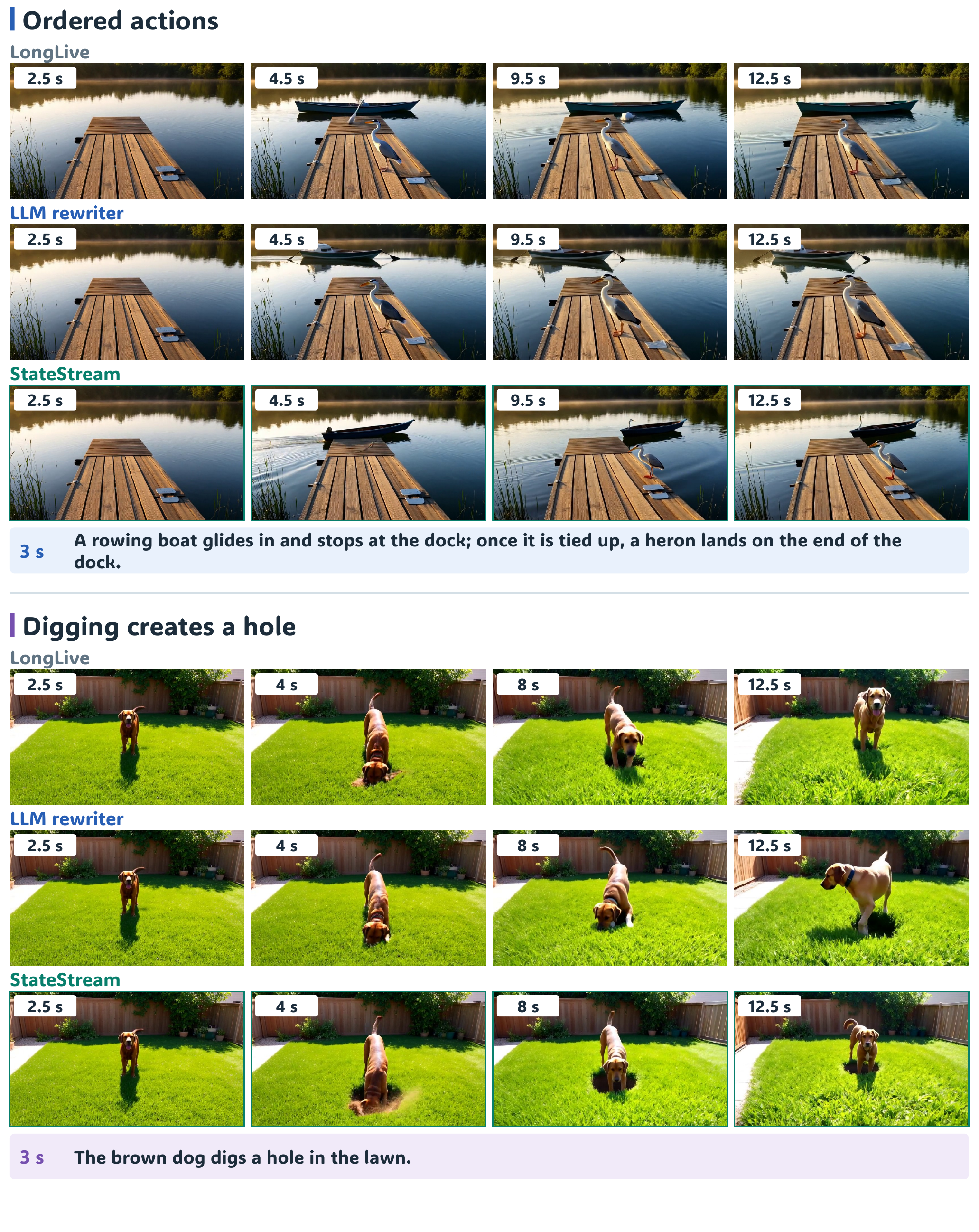}
\caption{Interaction execution and resulting state. Top: boat arrival precedes heron appearance in \method{}.}
\label{fig:supp_conditional_comparison}
\end{figure*}

\begin{figure*}[p]
\centering
\includegraphics[width=\textwidth,height=.93\textheight,keepaspectratio]{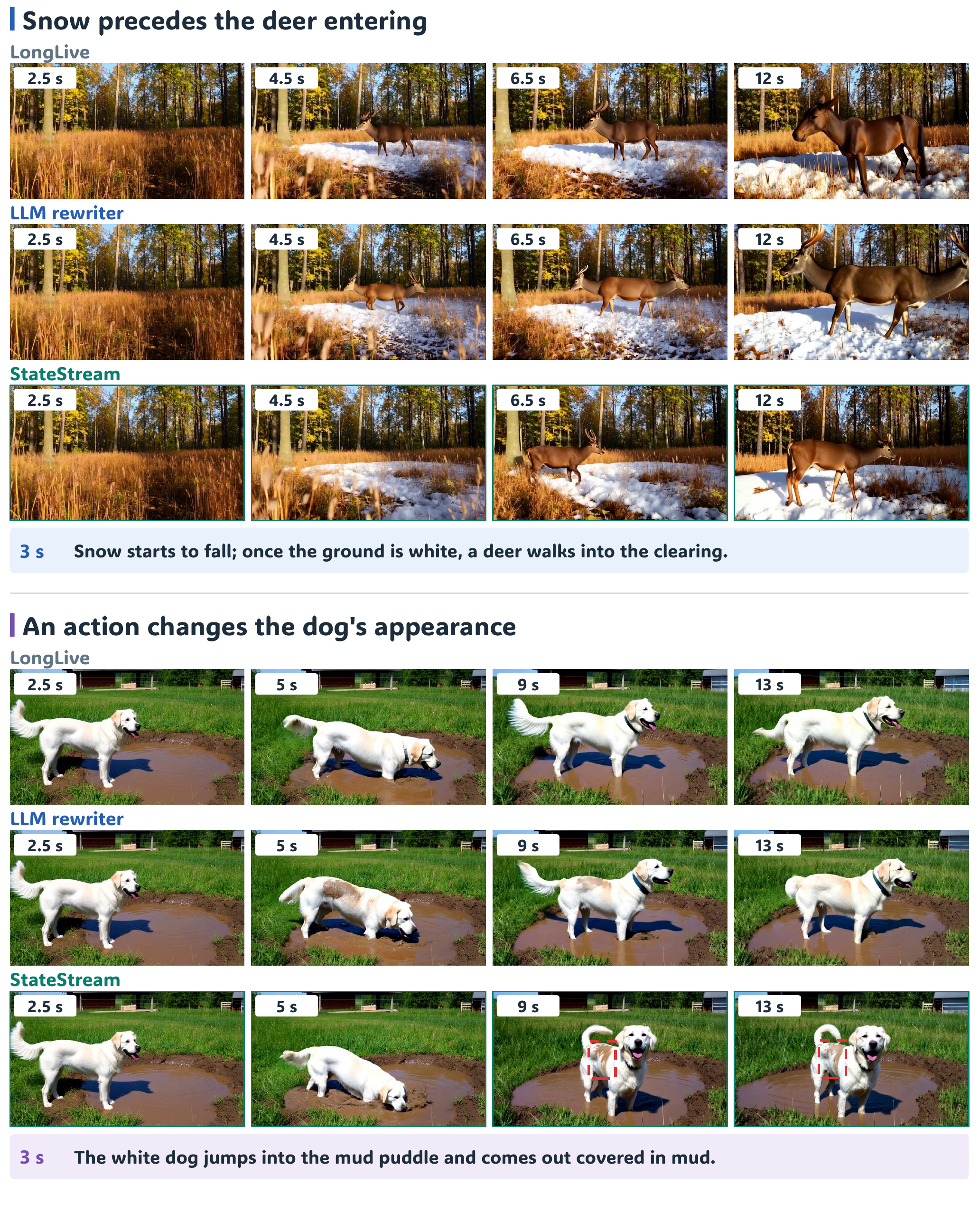}
\caption{Condition-dependent execution examples, the red dashed boxes show the better state control of 'a dog covered in mud'}
\label{fig:supp_condition_comparison}
\end{figure*}

\begin{figure*}[p]
\centering
\includegraphics[width=\textwidth,height=.93\textheight,keepaspectratio]{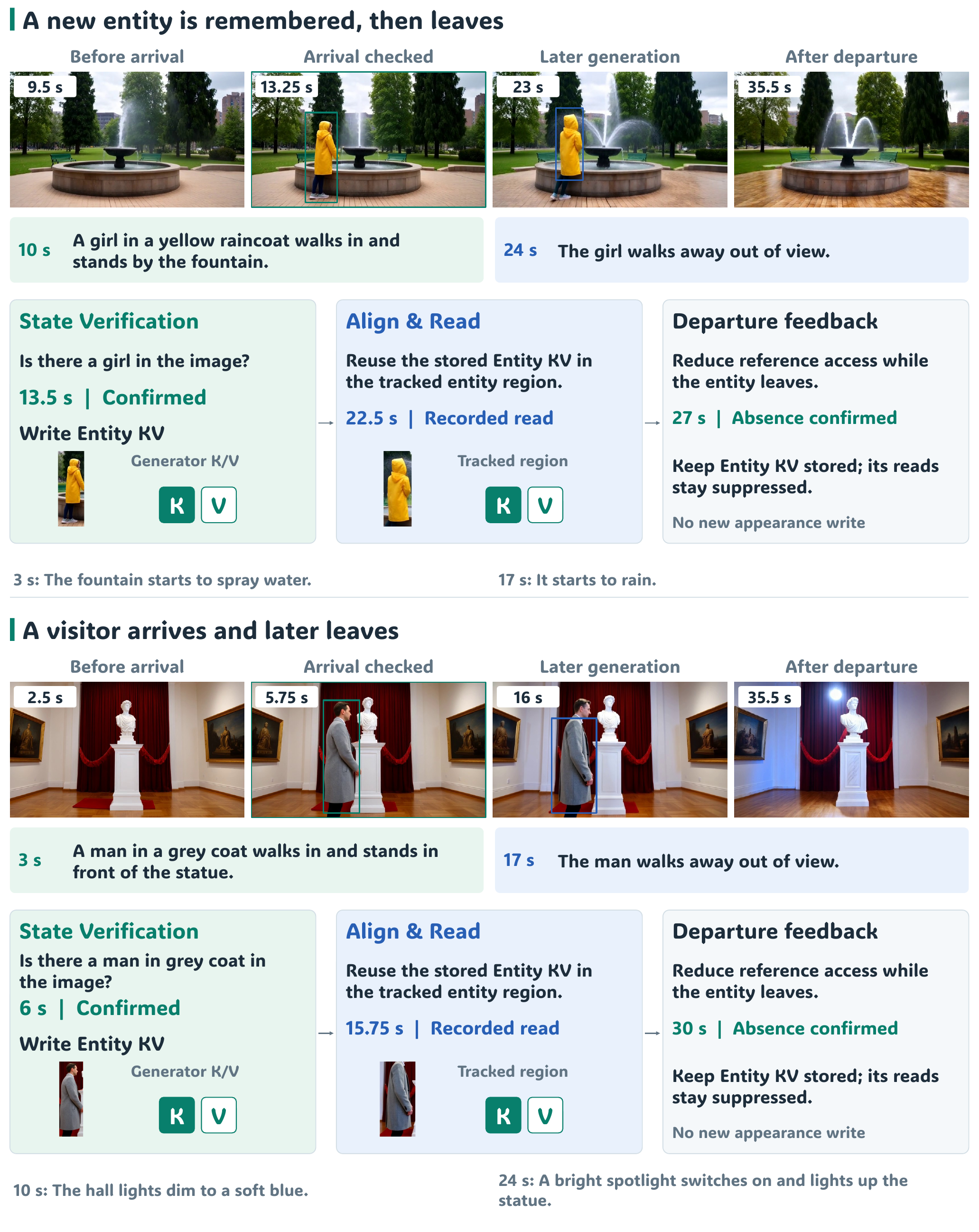}
\caption{Entity Memory through successive interactions. Confirmed arrival permits a local KV write; later generation reuses the reference in the tracked entity region. Departure reduces reference access, and confirmed absence completes the action while the stored entry remains suppressed.}
\label{fig:supp_verification_trace}
\end{figure*}

Figures~\ref{fig:supp_qualitative}, \ref{fig:supp_gallery_interactions}, and~\ref{fig:supp_gallery_extended} show additional interactions and 60\,s continuations with their input prompts. Figure~\ref{fig:supp_failures} shows remaining state loss, duplication, and drift. Comparisons use the same scenes, prompts, and frame times.
Figures~\ref{fig:supp_conditional_comparison}--\ref{fig:supp_condition_comparison}
compare dependent and conditional interactions, showing action timing
and resulting scene changes. Ordered interactions show intermediate
outcomes before the next action appears. These examples illustrate how execution plans
and visual checks coordinate interaction execution.
Figure~\ref{fig:supp_verification_trace} follows two entities from arrival
to departure using recorded execution and memory operations. Confirmed
presence permits an Entity KV write; later generation reuses that
reference within the tracked entity region. Departure reduces reference
access, and confirmed absence completes the action while the stored
entry remains suppressed. Visual confirmation thus controls local
writes and action completion, whereas entity state and region guide
reads. Insets show logged read-region bounds; decision times refer
to generation boundaries.

\section{Discussion and Future Work}
\label{sec:supp_analysis}
Section~\ref{subsec:limitation} discusses backbone dependence and the requirements for extending \method{} to stronger generators. Our current transfer evaluation covers related autoregressive models.
A broader direction is to develop interactive world models that retain the consequences of actions over long horizons. A future extension could maintain an estimate of observed world state alongside the state program, update it from visual evidence, and revise execution plans when observations differ from intended results. Combining this representation with learned state-transition dynamics could support predicting and comparing action outcomes before execution. The remaining failures in Fig.~\ref{fig:supp_failures} highlight the need for more reliable tracking and verification, memory that preserves local and global changes, and evaluation over longer continuous interactions.

\raggedbottom
\tcbset{selectedprompt/.style={before skip=4pt,after skip=4pt,top=1mm,bottom=1mm}}
\section{Agent Prompt}
\label{sec:supp_prompts}
The boxes reproduce the core agent instructions, with illustrative examples omitted from the shared instructions. Fields in braces are filled with the corresponding inputs; \texttt{command} refers to the interactive prompt.

\subsection{State program}
We place the shared instructions before each state-program prompt.

\begin{agentprompt}[selectedprompt]{Shared Instructions}
You are the world-state agent of an interactive video generator. You keep the state of the world
shown in the video as a small Python program. The program says only what exists and what state it is
in - no geometry, no pixels.
Statements you may use:
name = Thing("short visual description", at="where it is") # an object, plant, vehicle, part of a
  building ... in view
name = Creature("short visual description", at="where it is") # a person or an animal in view
name = Part("short visual description", of=other_name) # names a part that was already there (a
  car's hood)
name.<property> = "value" # a lasting change of that thing (e.g. door.state = "wide open",
  lamp.light = "on")
del name # the thing leaves the view or is destroyed for good
world.<property> = "value" # a scene-wide condition: sky, light, time_of_day, weather, fog, season
name.act("what it does", seconds=N) # a short motion that leaves no lasting change
verify("yes/no question about one frame", "yes" or "no") # how a viewer checks one lasting result on
  screen
Rules:
- every thing gets its own descriptive snake_case name (never x, obj, thing); never reuse a name for
  a different thing;
- every part the scene mentions separately is its own Thing (the door of a house, a straw in a
  drink, logs of a fire), so that it can change on its own later; never delete a thing because one
  of its parts changed;
- property names are real words (state, light, position, color, condition ...), never "attr";
- weather and light that fill the scene (fog, rain, snow, storm clouds, sunset, night) are always
  world.<property>, even when the command says what they cover;
- after the lasting lines of a command, one verify(...) per lasting result: a short yes/no question
  that a viewer can answer from a single frame, and the answer that is true once the result is
  there; no verify for a pure motion;
- write every lasting result the command causes, also the ones it does not spell out, as lines of
  the things involved (a full glass knocked off a table: the liquid is spilled on the floor; a glass
  vase dropped on stone breaks; snow left in hot sun melts into a puddle); a thing that ends up out
  of view is a del, never .at = "out of view"; nothing the command does not cause;
- "it", "the thing that ...", "what was put / left / taken ..." name a thing already in the program:
  use that name (e.g. crow.at = "on the red watering can");
- a command can depend on the world as it is now ("if it is raining, ...; otherwise ...", "the thing
  that fell", "what the wind blew off", "the bird that flew away"): look it up in the program so
  far, write only the lines of what actually happens, about the concrete thing the words refer to; a
  condition is only checked, never made true by your lines;
- no other statements, no imports, comments at most 8 words.
\end{agentprompt}

\begin{agentprompt}[selectedprompt]{State Update}
Current program:
```python
{program so far}
```
At {t} s the user types this command: "{interactive prompt}"
Write ONLY the new lines to append so that the program describes the world while the command happens
and after it:
- use the existing names; if the command changes a part that has no name yet (a car's hood, a ship's
  lights, a train's doors), first name it with Part(...); then one name.act(...) for the motion
  itself (shown only while it runs);
- then a Thing / Creature / property assignment / del for each lasting result (what is still true
  after it ends); a newcomer is a new Thing or Creature; something that leaves or is destroyed is a
  del (a thing that goes out of view is deleted, not moved to a place out of view with .at);
- if the command is only a motion that ends by itself (sways, trembles, flickers, drifts), write
  only the act line;
- never repeat unchanged lines.
Output only Python code.
\end{agentprompt}

\Needspace{10\baselineskip}
\subsection{Execution planning}
The State Planner uses this template to produce ordered steps and their visual checks (Sec.~\ref{sec:supp_planning}).

\begin{agentprompt}[selectedprompt]{Multi-Step Plan}
You control a video generator that renders a scene while a user gives commands. Reason about what
this command causes, then plan it as an ordered list of steps.

Scene now: "{opening description}"
Command: "{interactive prompt}" The code is what the world-state agent decided happens now: describe
only what the code changes, about the concrete things it names (the command may also mention things
that do not happen now, such as the branch of an "if" that does not apply); the sentence only says
what is seen, never why.
World-state change as code: {appended program lines}
Variables in this change: {variables}

Think step by step (silently) about the physical process: what happens first, what follows from it
(consequences the command does not spell out), and what the scene looks like when everything has
settled. Then answer with JSON only:
{"steps": [
{"sentence": "<ONE sentence describing the action while this step happens; name each object with its
  full description; for movement or leaving give the direction relative to the frame (left / right /
  top / bottom edge, toward or away from the camera)>",
"acts_on": [<variables from the list above that this step changes>],
"result": "<the visible state at the end of this step>",
"guard": {"question": "<yes/no question about ONE frame that is true once this step's result is
  visible>", "answer": "yes" or "no"},
"duration": <seconds until this step's result is complete, realistic for this object and action,
  0.75-8>}, ...],
"after": {"<variable>": "<one sentence for how it is once the action is done and it is still in
  view: how it looks and where it is; a person or animal also gets a small natural ongoing behaviour
  in place (never say it stays still); an object or a part is described only by its still state>",
  ...}}
Rules: 1 step if the command is a single simple change; 2-3 steps only when one thing causes a
change of ANOTHER thing (cause first, then each effect in order); never split one object's change
into stages (melting, opening, growing is one step); if the command says something happens when /
once / after something else is seen, that condition is its own earlier step whose guard checks the
condition, and the triggered action is the next step (it must not start before); every step must
leave a visible, lasting result (no flashes or sounds); a later step never describes its result
before it happens; guards ask about the step's own result, not about earlier steps; never mention
anything new that the command does not bring or imply; never write that something is absent or
missing except in a guard question; "after" only for variables still in view at the end. Every
sentence describes what is seen while it happens, in the present tense, as a caption of the video;
it is never an instruction to the video generator (not "Fade the sky ...", "Animate the leaves
...").
\end{agentprompt}



\clearpage

\clearpage
\clearpage
\clearpage

\clearpage


\clearpage

\clearpage

\end{document}